\documentclass{article}

\usepackage{PRIMEarxiv}

\usepackage[utf8]{inputenc}
\usepackage[T1]{fontenc}
\usepackage{hyperref}
\usepackage{url}
\usepackage{booktabs}
\usepackage{amsmath}
\usepackage{amsfonts}
\usepackage{amssymb}
\usepackage{nicefrac}
\usepackage{microtype}
\usepackage{graphicx}
\usepackage{subcaption}
\usepackage{float}
\usepackage{placeins}
\graphicspath{{./}{3D_Grounding_Detection/images/}}

\newcommand{\method}{CT-3GDINO}

\title{Pseudo-Text-Conditioned 3D Grounding DINO for Organ Localization in Abdominal CT}
\author{
Siqi Chen$^{1}$, Han Gong$^{1}$, Keyi Hou$^{1}$, Jingxuan Yang$^{1}$, Sheethal Bhat$^{1}$, Andreas Maier$^{1}$\\
\vspace{0.2cm}
\small $^{1}$Friedrich-Alexander-Universit\"at Erlangen-N\"urnberg, Germany\\
\small \texttt{\{siqi.chen, han.gong, keyi.hou, jingxuan.yang, sheethal.bhat, andreas.maier\}@fau.de}
}

\begin{document}
\maketitle

\begin{abstract}
Reliable organ localization in abdominal CT can provide spatial priors for downstream trauma analysis. We propose \method, a lightweight 3D detector that adapts a Grounding-DINO-style query-based architecture to fixed organ localization using frozen pseudo-text class tokens instead of a real text encoder. The model combines a Swin3D visual backbone, bidirectional feature enhancement, pseudo-text-guided query selection, and a cross-modality decoder to predict normalized 3D boxes for liver, spleen, left kidney, right kidney, and bowel. We train and evaluate on 193 matched RSNA/RATIC CT volumes with segmentation-derived boxes. The best multi-scale model, trained from scratch, achieves 0.5830 overall top-1 class-wise mAP over 3D IoU thresholds from 0.1 to 0.7, outperforming fixed- and trainable-backbone classification-pretrained variants with 0.5570 and 0.4657 mAP. Performance is strong for coarse localization, with 0.9649 AP at IoU 0.1, but remains limited for strict box alignment, with 0.1552 AP at IoU 0.7. These results establish \method\ as an open-source baseline for pseudo-text-conditioned 3D organ localization and motivate future work on localization-aware pretraining, richer multimodal conditioning, and injury-focused detection.
\end{abstract}

\section{Introduction}

Rapid and accurate identification of traumatic abdominal injuries on CT is critical for timely intervention in emergency care. Abdominal CT provides volumetric evidence about the location and severity of injuries, but downstream injury classification becomes more reliable when the relevant anatomical structures are first localized consistently across studies. In this work, we focus on organ-level 3D bounding-box localization as a structured intermediate task for downstream traumatic injury analysis. The present model predicts organ boxes that can later serve as spatial priors for injury-focused models.

Recent query-based detectors such as DETR and DINO formulate object detection as set prediction and reduce reliance on hand-designed anchors and non-maximum suppression \cite{carion2020end,zhang2022dino}. Grounding DINO further introduces grounded vision-language pretraining, cross-modal fusion between visual and textual features, text-guided query selection, and multi-scale deformable attention inherited from Deformable DETR, which together improve class-specific feature encoding and convergence in the 2D setting \cite{zhu2020deformable,liu2024grounding}. Recent medical Grounding-DINO-style and exemplar-conditioned methods have also shown state-of-the-art or competitive performance in medical detection, suggesting that semantic or prototype guidance can extract clinically useful features when annotations are limited \cite{bhat2025exemplar,lin2024ct}. These results motivate us to adapt the Grounding DINO paradigm to 3D volumetric detection while preserving a semantic-conditioning interface for future language prompts, exemplar prompts, and few-shot class extension. However, the 3D setting remains non-trivial because volumetric feature extraction is substantially more memory demanding than 2D feature extraction, organ box annotations are limited, and the fixed organ vocabulary used in this first study favors a lightweight semantic branch.

We propose \method, a 3D pseudo-text-conditioned detector for abdominal organ localization. \method\ uses frozen Xavier-initialized pseudo text class embeddings and a lightweight trainable projection MLP to provide class anchors for liver, spleen, left kidney, right kidney, and bowel. CT volumes are processed by a Swin3D backbone, multimodal features are refined through a bidirectional feature enhancer, and language-guided query selection initializes object queries from semantically relevant image tokens. A cross-modality decoder then attends to both image and pseudo-text features and predicts normalized 3D boxes of the form $(c_x,c_y,c_z,w,h,d)$. This paper presents a first 3D semantic-conditioned Grounding DINO baseline evaluated on CT.

The remainder of this paper presents the complete pipeline, describes the implemented detector components, and reports the final quantitative and qualitative results. Detailed descriptions of the supporting backbone, feature-enhancement, and query-selection components are provided in Appendix~\ref{app:backbone}, Appendix~\ref{app:feature_enhancer}, and Appendix~\ref{app:query_selection}, respectively.

The main contributions are:
\begin{itemize}
    \item We formulate abdominal organ localization as semantic-conditioned 3D set prediction using pseudo text class tokens as the semantic interface.
    \item We implement the pseudo text generator, cross-modality decoder, prediction heads, Hungarian matching loss, 3D IoU/mAP evaluation, and the end-to-end training and evaluation pipeline.
    \item We evaluate multi-scale variants with and without classification-pretrained Swin3D backbones and clarify that the current classification pretraining initializes only the visual feature extractor.
    \item We present \method\ as an open-source baseline for 3D multimodal detection with minimal annotations and discuss future extensions to language-pretrained, exemplar-conditioned, and few-shot 3D detection.
\end{itemize}

\section{Related Work}

\paragraph{Query-based object detection.}
DETR introduced end-to-end object detection as a bipartite set prediction problem, replacing anchor assignment and post-processing heuristics with a transformer decoder and Hungarian matching \cite{carion2020end}. Deformable DETR improved convergence by attending to a sparse set of sampling locations on multi-scale feature maps, and DINO further improved DETR-style optimization through denoising and stronger query formulation \cite{zhu2020deformable,zhang2022dino}. These methods provide the foundation for our detector because organ localization can naturally be represented as a small set of class-conditioned boxes.

\paragraph{Grounded and semantic-conditioned detection.}
Grounding DINO combines DINO-style detection with grounded pretraining, cross-modal fusion between image and text features, language-guided query selection, and multi-scale deformable attention heads \cite{liu2024grounding}. This combination is strong because text-image cross-attention helps encode class-specific visual features, grounded pretraining provides transferable semantic representations, and multi-scale deformable attention accelerates convergence by focusing attention on a small number of relevant feature locations. Recent medical Grounding-DINO-style and exemplar-conditioned methods have also shown strong, state-of-the-art or competitive results in medical detection, including generalized lesion detection with exemplar guidance \cite{bhat2025exemplar,lin2024ct}. We therefore choose a Grounding-DINO-inspired formulation with a semantic-conditioning interface; the long-term goal is a 3D detector that can use language, prototypes, or exemplars to detect new classes with fewer annotations. Our current work follows this semantic-conditioning principle and uses pseudo class tokens for the fixed target vocabulary.

\paragraph{3D visual backbones for medical volumes.}
Transformer backbones are attractive for volumetric data because self-attention can capture long-range anatomical context. However, global attention over 3D volumes is memory intensive. Swin-style hierarchical window attention reduces this cost by applying attention within local windows while preserving multi-scale features \cite{yang2025swin3d}. We adopt a Swin3D-style visual encoder for CT feature extraction and use multi-scale outputs for downstream detection.

\paragraph{Organ localization from segmentation-derived boxes.}
Dense organ segmentations provide precise supervision for medical image analysis, as reflected by large-scale segmentation benchmarks and multi-organ CT segmentation tools \cite{antonelli2022medical,wasserthal2023totalsegmentator}. However, full voxel-level labels are expensive to create and remain scarce in trauma-specific datasets \cite{rajchl2016deepcut,hermans2024rsna,rudie2024rsna}. Bounding boxes offer a lighter representation: they retain the anatomical extent needed to crop or attend to organs with no voxel-level prediction at inference time, and box-level supervision has been used as a practical alternative to dense labels in medical image analysis and CT lesion detection \cite{rajchl2016deepcut,yan2018deeplesion}. In this study, segmentation masks are used only to generate supervision boxes. The detector itself predicts boxes directly from CT volumes, which makes the task closer to downstream triage systems that need organ regions for injury classification, report generation, or focused human review.

\section{Method}

\subsection{Overview}

\begin{figure*}[t]
    \centering
    \includegraphics[width=\linewidth]{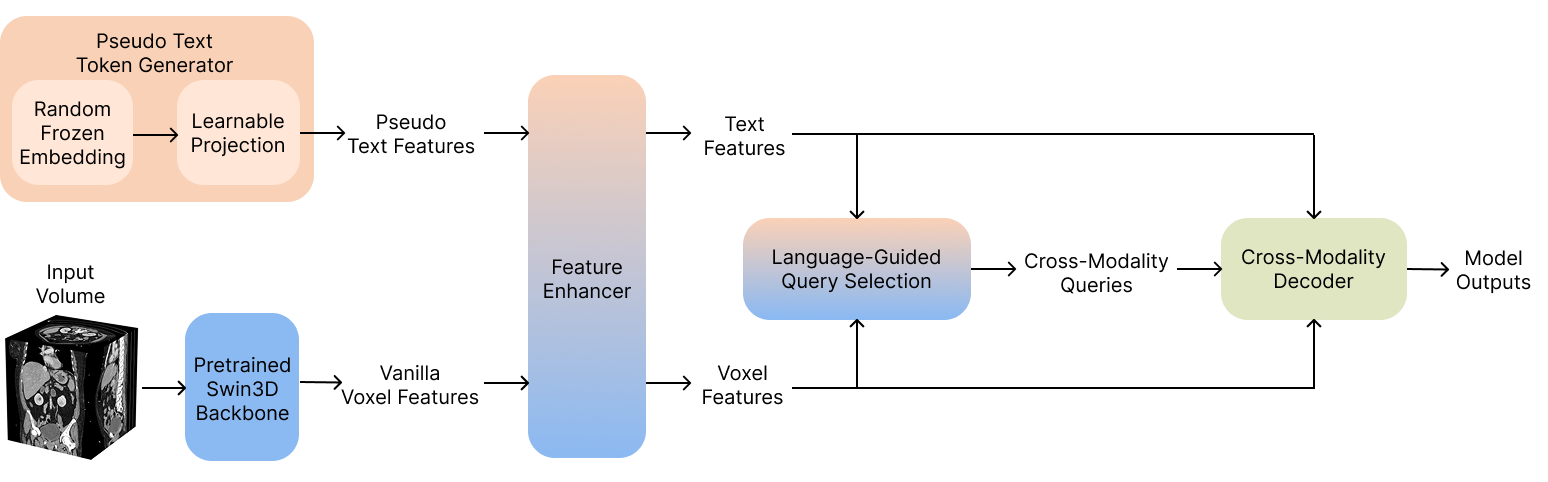}
    \caption{Overview of \method, a 3D Grounding-DINO-style detector evaluated on abdominal CT scans. CT volumes are processed by a trainable Swin3D backbone to obtain multi-scale volumetric features. The text branch is simulated using frozen random class-wise pseudo text embeddings for the five organs, followed by a trainable projection MLP. A bidirectional feature enhancer performs cross-attention between 3D image tokens and pseudo text tokens, language-guided top-$K$ query selection orders candidate hypothesis regions, and the cross-modality decoder outputs class-wise volumetric box predictions.}
    \label{fig:pipeline}
\end{figure*}

Figure~\ref{fig:pipeline} summarizes the proposed pipeline. Given an abdominal CT volume $V \in \mathbb{R}^{D \times H \times W}$, the model first extracts multi-scale visual features with a 3D Swin backbone. In parallel, pseudo text class tokens represent the five organ categories. The feature enhancer performs early bidirectional interaction between visual and pseudo-language tokens. Language-guided query selection then chooses image tokens with high semantic relevance as initial object queries. Finally, the cross-modality decoder refines the queries by attending to both modalities and outputs class logits and normalized 3D boxes.

Let the backbone output a set of selected multi-scale feature maps $\{F^{(l)}\}_{l=1}^{L}$. Each feature map is projected to a common hidden dimension $d$ and flattened into tokens. Concatenating all selected scales gives image tokens
\begin{equation}
X^{I} \in \mathbb{R}^{B \times N_I \times d},
\end{equation}
where $B$ is the batch size and $N_I$ is the number of visual tokens. The pseudo text branch generates class tokens
\begin{equation}
X^{T} \in \mathbb{R}^{B \times C \times d},
\end{equation}
where $C=5$ for liver, spleen, left kidney, right kidney, and bowel. The final model predicts $N_q$ query outputs, each with a class distribution over $C+1$ categories including background and a normalized 3D box.

The overall design deliberately separates three forms of information. The Swin3D backbone supplies spatial and anatomical evidence from the CT volume. The pseudo text branch supplies class identity and allows the architecture to keep the same semantic-conditioning interface as a grounded detector. The query pathway bridges these representations by selecting candidate spatial tokens and refining them into organ-level predictions. This modularity is useful for collaboration: the backbone, feature enhancer, and query selection modules can be improved independently, while the detector interface and training objective remain unchanged.

\subsection{Swin3D Visual Encoder}

The visual encoder converts a CT volume into hierarchical volumetric representations. The input is partitioned into non-overlapping 3D patches and projected into an embedding space. Subsequent transformer stages use window-based self-attention and patch merging to enlarge the receptive field while reducing spatial resolution. The detector uses multi-scale outputs from selected later stages, which preserves coarse anatomical context and higher-resolution spatial information. The full backbone implementation and ablation details are summarized in Appendix~\ref{app:backbone}.

\subsection{Pseudo Text Token Generator}
\label{subsec:pseudo_text}

The original Grounding DINO uses a text encoder to process free-form prompts. For the present organ localization task, the vocabulary is fixed and small, so we use a pseudo text token generator. This module takes the fixed list of organ class indices as input. Let $E \in \mathbb{R}^{C \times d}$ be a class embedding matrix with one row per organ class. At initialization, $E$ is sampled once using Xavier uniform initialization and is then frozen. A trainable two-layer projection MLP maps these frozen random class anchors into the detector hidden space:
\begin{equation}
X^{T}_0 = \phi(E), \qquad \phi(E)=W_2\,\sigma(W_1E+b_1)+b_2,
\end{equation}
where $\sigma$ denotes ReLU. The output is a pseudo text token tensor $X^{T}_0 \in \mathbb{R}^{B \times C \times d}$ after expansion along the batch dimension. Thus the input is a fixed organ vocabulary and the output is one hidden-dimensional class token per organ for every batch element. The embedding table remains frozen, while the projection MLP is optimized jointly with the detector. This keeps the semantic branch lightweight while still providing class-specific anchors for subsequent multimodal attention.

The pseudo tokens are compact class-conditioned parameters that allow the detector to preserve the interface of a grounded model without requiring large-scale 3D language-image pretraining. This design is practical for a fixed organ vocabulary and provides a controlled baseline for future replacement with richer report- or prompt-based text features.

This component is also the main architectural difference between our framework and a conventional closed-set 3D detector. A standard detector would use only a classification head to associate each query with an organ class after decoding. \method\ injects class information before query selection and decoding. The pseudo-language features enable early class-conditioned feature interaction and make the decoder queries aware of the target organ set throughout the prediction process.

\subsection{Bidirectional Feature Enhancer}

Before query selection and decoding, image and pseudo text tokens are jointly refined by a bidirectional feature enhancer. Each layer applies self-attention within each modality, image-to-text cross-attention, text-to-image cross-attention, feed-forward networks, residual connections, and layer normalization. This module transforms appearance-driven visual tokens into category-aware visual representations and grounds pseudo text tokens in image evidence. The complete description and ablation are provided in Appendix~\ref{app:feature_enhancer}.

\subsection{Language-Guided Query Selection}

After feature enhancement, the model initializes decoder queries from image tokens that are most relevant to the pseudo text tokens. Image and text features are projected to a shared hidden dimension and their dot-product similarity is computed as
\begin{equation}
S_{b,i,c}=\langle \bar{X}^{I}_{b,i}, \bar{X}^{T}_{b,c}\rangle.
\end{equation}
Each image token receives a semantic relevance score by taking the maximum similarity over class tokens:
\begin{equation}
s_{b,i}=\max_{c} S_{b,i,c}.
\end{equation}
The top-$K$ image tokens are selected as query initialization features, where $K=N_q$. These selected features are combined with learnable content queries before entering the decoder. Appendix~\ref{app:query_selection} summarizes the full query-selection report and its ablations.

This strategy is particularly suitable for organ localization because the number of foreground objects is small compared with the number of volumetric tokens. Most CT tokens correspond to background tissue, air, bone, or non-target anatomy. Selecting the top semantic tokens before decoding reduces the burden on the decoder by starting from image locations that already have high pseudo-text similarity.

\subsection{Cross-Modality Decoder and Prediction Heads}
\label{subsec:decoder}

The cross-modality decoder is responsible for transforming initialized queries into object-level predictions. Each decoder layer contains four operations: query self-attention, image cross-attention, text cross-attention, and a feed-forward network. With query tokens $Q$, enhanced image tokens $\hat{X}^{I}$, and enhanced pseudo text tokens $\hat{X}^{T}$, one decoder layer can be summarized as
\begin{align}
Q_1 &= \operatorname{LN}\bigl(Q + \operatorname{MSA}(Q,Q,Q)\bigr), \\
Q_2 &= \operatorname{LN}\bigl(Q_1 + \operatorname{CA}(Q_1,\hat{X}^{I},\hat{X}^{I})\bigr), \\
Q_3 &= \operatorname{LN}\bigl(Q_2 + \operatorname{CA}(Q_2,\hat{X}^{T},\hat{X}^{T})\bigr), \\
Q_{\mathrm{out}} &= \operatorname{LN}\bigl(Q_3 + \operatorname{FFN}(Q_3)\bigr).
\end{align}
Here, $\operatorname{MSA}$ denotes multi-head self-attention, $\operatorname{CA}$ denotes multi-head cross-attention, and $\operatorname{LN}$ denotes layer normalization. The implementation uses six decoder layers, eight attention heads, hidden dimension $d=256$, and feed-forward dimension 2048.

The final query embeddings are passed to two prediction heads. A linear classifier outputs logits in $\mathbb{R}^{C+1}$, where the additional class represents background. A three-layer MLP predicts six normalized box coordinates:
\begin{equation}
\hat{b}=(\hat{c}_x,\hat{c}_y,\hat{c}_z,\hat{w},\hat{h},\hat{d}) \in [0,1]^6.
\end{equation}
A sigmoid activation constrains the box parameters to the normalized volume coordinate system.

The decoder alternates between spatial evidence and semantic evidence. Image cross-attention lets object queries collect volumetric information from the enhanced CT tokens, which is necessary for locating organ centers and estimating box extents. Text cross-attention reintroduces class-conditioned information after spatial aggregation, reducing the risk that queries collapse into purely appearance-based proposals. The final classifier and regressor are intentionally lightweight so that most modeling capacity remains in the shared multimodal representations, with minimal class-specific prediction heads.

In the original 2D Grounding DINO setting, multi-scale deformable attention is an important part of efficient high-performance detection because each query samples only a sparse set of relevant points across feature levels. Translating this head directly to 3D would require sparse sampling over depth, height, width, and scale. The present implementation is therefore a first 3D adaptation that uses multi-scale Swin3D features and dense cross-attention; a true 3D multi-scale deformable attention decoder is a natural next step for faster convergence and more precise box refinement.

\subsection{Set-Based Training Objective}

Following DETR-style detectors, training uses one-to-one bipartite matching between predictions and ground-truth boxes. For a prediction $i$ and target $j$, the matching cost combines classification probability, L1 box distance, and 3D generalized IoU:
\begin{equation}
\mathcal{C}_{ij}=\lambda_{\mathrm{cls}}\mathcal{C}^{\mathrm{cls}}_{ij}
+\lambda_{\ell_1}\lVert \hat{b}_i-b_j\rVert_1
+\lambda_{\mathrm{giou}}\mathcal{C}^{\mathrm{giou}}_{ij}.
\end{equation}
The Hungarian algorithm finds the minimum-cost assignment. The final training loss uses the same components after matching:
\begin{equation}
\mathcal{L}=\lambda_{\mathrm{ce}}\mathcal{L}_{\mathrm{ce}}
+\lambda_{\ell_1}\mathcal{L}_{\ell_1}
+\lambda_{\mathrm{giou}}\mathcal{L}_{\mathrm{giou}}.
\end{equation}
The background class weight is reduced to limit the effect of unmatched queries. This objective allows the model to train without anchor boxes or non-maximum suppression, which is suitable for organ localization where each class appears at most once per study.

For 3D boxes, IoU and generalized IoU are computed from axis-aligned cuboids. If a box is represented by center and size, it is first converted to minimum and maximum corners along the $x$, $y$, and $z$ axes. The intersection volume is the product of positive overlaps along the three axes, and the union is the sum of both box volumes minus the intersection. The generalized IoU term adds a penalty based on the smallest enclosing cuboid, which provides a useful training signal even when predicted and target boxes do not overlap.

\section{Data and Experimental Setup}

\subsection{Dataset and Box Generation}

We evaluate on the RSNA 2023 abdominal trauma dataset and the corresponding RSNA Abdominal Traumatic Injury CT (RATIC) public release \cite{hermans2024rsna,rudie2024rsna}. RATIC contains adult abdominal and pelvic trauma CT scans collected from 4,274 patients, with 6,481 image series contributed by 23 institutions across 14 countries and six continents. The dataset provides multi-level annotations for traumatic abdominal injury, including study-level labels for solid-organ injuries, image-level labels for bowel/mesenteric injury and active extravasation, and voxelwise organ segmentations for a subset of 206 training series.

\FloatBarrier
\begin{figure}[!htbp]
    \centering
    \includegraphics[width=0.95\linewidth]{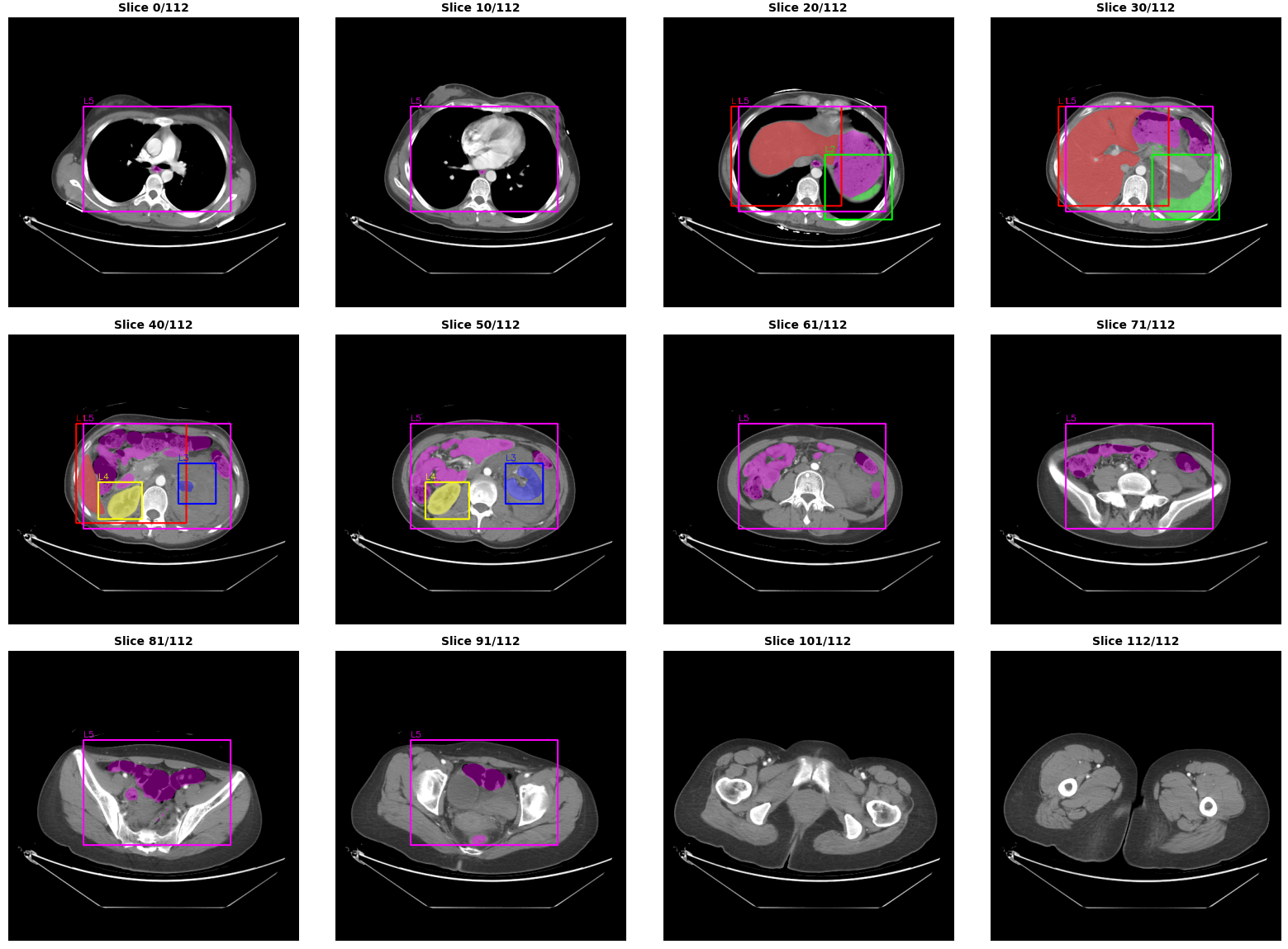}
    \caption{Visualization of the segmentation-derived localization targets used for box generation. Organ colors are consistent throughout the paper: liver is shown in red, spleen in green, left kidney in blue, right kidney in yellow, and bowel in magenta. The overlays illustrate how dense organ masks provide spatial supervision from which enclosing 3D boxes are extracted. Because the RSNA segmentations were generated through an nnU-Net-based pipeline followed by manual correction, the masks are useful but may still contain residual errors; the resulting boxes should therefore be treated as noisy segmentation-derived ground truth with residual boundary uncertainty.}
    \label{fig:dataset_visualization}
\end{figure}
\FloatBarrier

Only the segmentation subset is used in this work because our goal is organ localization. This means that dense spatial supervision is available for only 206 of 6,481 public image series, i.e., approximately 3.2\% of the series, or 4.8\% relative to the 4,274 patients. After preprocessing and filtering, our experiments use 193 volumes split into 135 training, 28 validation, and 30 test volumes. Since each volume contains at most one box per target organ, this corresponds to at most 135/28/30 boxes per organ in the train/validation/test splits before missing-label effects. The RATIC dataset description states that voxelwise segmentations were manually corrected after training an nnU-Net on the TotalSegmentator dataset for the challenge organs: liver, spleen, left kidney, right kidney, and bowel \cite{rudie2024rsna}. The bowel label represents a combined gastrointestinal tract label that includes esophagus, stomach, duodenum, small bowel, and colon. Therefore, the extracted boxes are segmentation-derived supervision with residual uncertainty relative to manual box annotations. Residual segmentation errors, bowel-label aggregation, scan-coverage differences, and interpolation during preprocessing may propagate to the generated boxes.

For each connected organ region, the minimum and maximum foreground coordinates are converted to a center-size box. Multiple connected components of the same label are merged into one class-level box, matching the top-1 class-wise evaluation setting. Boxes are normalized by the resized volume dimensions and represented as $(c_x,c_y,c_z,w,h,d)$.

The five localization labels are liver, spleen, left kidney, right kidney, and bowel. CT intensities are windowed and normalized to $[0,1]$. Volumes are resized proportionally according to a target width, and the corresponding masks are resized with nearest-neighbor interpolation before box extraction. Training uses 3D augmentations including rotations, scaling, elastic deformation, and intensity jitter when enabled.

The preprocessing pipeline is designed to avoid manual box annotation. Each segmentation mask is treated as an intermediate annotation source: after resampling and resizing, the foreground coordinates of each organ label are converted into a single enclosing cuboid. This choice intentionally ignores fine organ shape and focuses the learning target on spatial extent. It also makes the annotations compatible with set-prediction detection losses, which use a small list of boxes and class labels per volume. Figure~\ref{fig:dataset_visualization} illustrates the segmentation-derived supervision and the class-color convention used in our visual analysis.

Because CT studies can have different numbers of slices and different in-plane resolutions, proportional resizing preserves approximate anatomical proportions without forcing all axes to a fixed cubic shape. The target width controls memory usage, while the remaining dimensions are scaled by the same factor. During batching, volumes are padded to the maximum size in the batch, and the target boxes remain normalized, so the loss is independent of the padded voxel dimensions.

\subsection{Training Protocol}

The detector is implemented in PyTorch. Unless otherwise stated, the model uses 100 object queries, hidden dimension 256, six feature-enhancer layers, six decoder layers, and eight attention heads. Optimization uses AdamW with weight decay, a linear warm-up phase, cosine learning-rate decay, and gradient clipping. The experiments reported in the final comparison are trained for 500 epochs with a batch size of 2.

We compare three multi-scale settings: (1) a model trained without backbone pretraining, (2) a model initialized from a patient-level injury classification-pretrained backbone and fine-tuned end-to-end, and (3) a model using the same pretrained backbone while keeping the backbone fixed. In the pretrained variants, only the Swin3D visual feature extractor is initialized from the classification checkpoint; the pseudo token generator, feature enhancer, query selection module, decoder, classifier, and box regressor are trained from scratch. This comparison tests whether single-scale classification pretraining transfers to multi-scale 3D localization.

The pretraining task uses patient-level abdominal trauma labels and therefore encourages the backbone to encode global injury-related cues. This is a useful initialization hypothesis, but it may mismatch the localization objective because classification pretraining omits the final multi-scale feature hierarchy, semantic query selection, and 3D box regression losses. In the detection task, gradients must preserve spatial information because the model must regress organ centers and dimensions. The frozen-backbone variant tests whether pretrained features are already sufficient, while the trainable-backbone variant tests whether fine-tuning can adapt those features. The model without pretraining serves as a task-specific baseline in which all visual features are learned directly from the detection loss.

\subsection{Evaluation Metric}

Evaluation uses class-wise average precision under 3D IoU thresholds from 0.1 to 0.7. For each class, predictions are sorted by confidence and greedily matched to ground-truth boxes at the selected IoU threshold. Because each organ class has at most one target instance per study, we report top-1 class-wise organ localization. The overall mAP is the mean across evaluated IoU thresholds:
\begin{equation}
\mathrm{mAP}=\frac{1}{T}\sum_{t=1}^{T}\mathrm{mAP}(\tau_t), \qquad \tau_t \in \{0.1,0.2,\ldots,0.7\}.
\end{equation}

Reporting a range of IoU thresholds is important for 3D medical localization. Low thresholds such as 0.1 and 0.2 measure whether the model finds the correct anatomical region, which is useful for coarse triage or initializing downstream crops. Higher thresholds such as 0.5 and 0.7 measure whether the predicted cuboid tightly matches the segmentation-derived extent. The strict thresholds are more sensitive to small boundary errors because a few voxels of mismatch along each axis can compound into a large 3D volume difference.

\section{Results}

\subsection{Final Quantitative Performance}
\label{sec:final_quantitative}

\begin{figure}[!htbp]
    \centering
    \includegraphics[width=\linewidth]{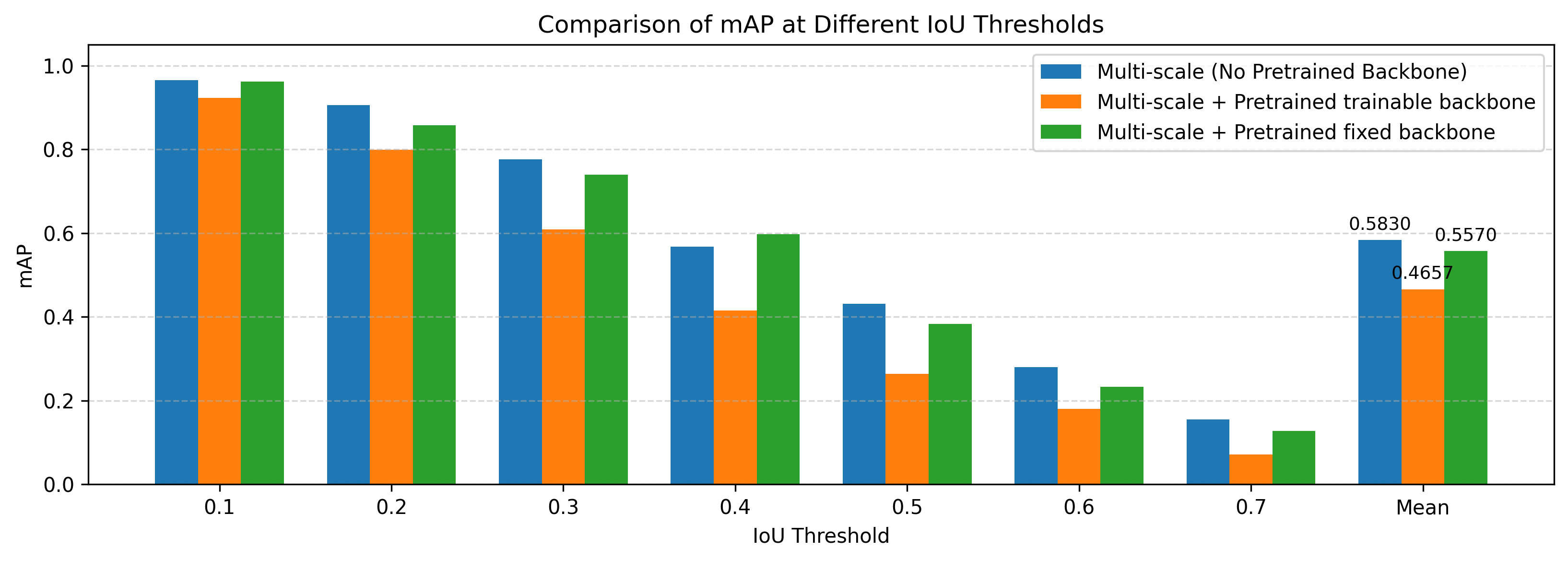}
    \caption{Top-1 class-wise 3D bounding-box AP across IoU thresholds for three multi-scale variants. ``Multi-scale, no pretrained backbone'' uses multi-scale Swin3D features trained from scratch together with the detector. ``Fixed pretrained'' and ``trainable pretrained'' initialize only the Swin3D feature extractor from a patient-level classification checkpoint, with the remaining detection modules trained from scratch. The scratch multi-scale variant achieves the highest overall mAP of 0.5830, suggesting that task-aligned multi-scale localization training is more useful here than classification pretraining alone.}
    \label{fig:map_grouped}
\end{figure}

Figure~\ref{fig:map_grouped} summarizes the final pretraining ablation. The multi-scale model trained without a pretrained backbone achieves the best overall mAP of 0.5830. It outperforms the fixed-backbone pretrained variant with 0.5570 mAP and the trainable-backbone pretrained variant with 0.4657 mAP. The best model obtains strong AP at loose and moderate thresholds, but performance decreases at stricter thresholds, indicating that coarse anatomical localization is easier than tight 3D boundary alignment.

\FloatBarrier
\FloatBarrier

The result suggests that task alignment is more important than pretraining alone in the current setting. The pretrained checkpoint was optimized for patient-level injury classification using the visual feature extractor, whereas the final detector requires spatially precise, multi-scale features for 3D box regression and semantic query selection. Freezing the backbone preserves the classification representation but prevents adaptation to localization. Fine-tuning provides more flexibility, but the initialization may still bias optimization toward global classification features and away from scale-aware organ boundaries. Training the backbone directly within the final detection objective allows the visual encoder, feature enhancer, query selection, and decoder to co-adapt from the beginning. This explains why the scratch multi-scale variant can outperform the pretrained variants despite the general expectation that pretraining is beneficial.

Figure~\ref{fig:map_grouped} further shows that all variants follow the same general trend: performance is high at lower IoU thresholds and decreases as the overlap criterion becomes stricter. This indicates that the models generally learn where the organs are located, but exact 3D extents remain difficult. The decline is expected because box errors accumulate along three axes; a prediction that appears reasonable on a representative 2D slice may still lose substantial IoU if its depth extent is inaccurate. The results therefore support using both quantitative AP and qualitative slice overlays in the final paper.

\subsection{Component Ablations}

Module-specific analyses further clarify the behavior of the full system. The Swin3D backbone ablation shows that window attention is necessary for practical volumetric training: full attention already requires 31.99 GB at target width 64, whereas window attention reduces this to 4.91 GB and allows experiments at target widths 96 and 128. This memory reduction is important because higher spatial resolution is one of the most direct ways to improve strict-IoU localization.

The feature-enhancer ablation shows that early bidirectional multimodal interaction improves detection quality. The full feature enhancer increases overall mAP from 0.4602 to 0.5570 compared with the model without the module. The improvement is particularly large at intermediate thresholds, suggesting that pseudo-text-conditioned visual refinement helps both semantic discrimination and spatial alignment.

The language-guided query-selection ablation shows that semantically informed query initialization is beneficial for 3D organ localization. Compared with purely learnable queries and random visual-token initialization, language-guided selection achieves the strongest module-level mean AP in the reported ablation. Its gains are especially visible at stricter IoU thresholds, which suggests that selecting semantically relevant visual tokens helps the decoder refine organ boundaries more effectively. Detailed module analyses are provided in Appendix~\ref{app:backbone}, Appendix~\ref{app:feature_enhancer}, and Appendix~\ref{app:query_selection}.

\subsection{Qualitative Localization Samples}

\begin{figure}[!t]
    \centering
    \includegraphics[width=0.96\linewidth]{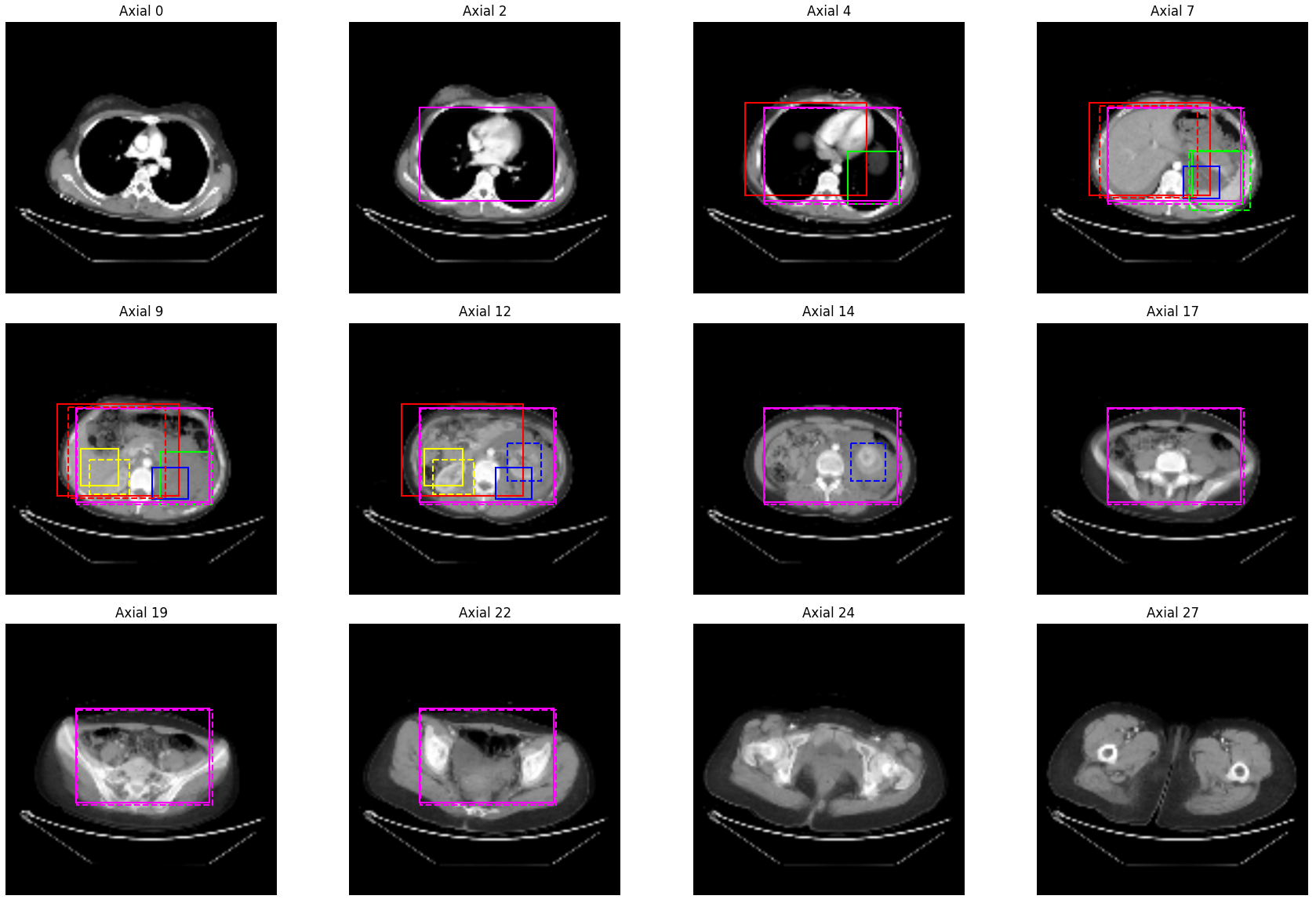}
    \caption{Qualitative localization samples for \method. Dashed boxes are segmentation-derived ground-truth boxes and solid boxes are predicted boxes. Red denotes liver, green denotes spleen, blue denotes left kidney, yellow denotes right kidney, and magenta denotes bowel. These examples show that the model generally identifies the correct abdominal regions, while differences between dashed and solid boxes reveal common 3D localization errors such as over-extended depth, loose boundaries, or imperfect alignment on compact organs. Because the ground-truth boxes are derived from segmentation masks that were generated by an external segmentation pipeline and then corrected, part of the apparent mismatch can also reflect noise in the segmentation-derived supervision.}
    \label{fig:qualitative}
\end{figure}

Qualitative output examples are included because they show whether the predicted boxes are anatomically plausible and where failures occur. Figure~\ref{fig:qualitative} compares the top-1 predicted boxes with segmentation-derived ground-truth boxes on representative CT slices. The color coding follows the dataset visualization: liver is red, spleen is green, left kidney is blue, right kidney is yellow, and bowel is magenta. Dashed boxes denote ground truth and solid boxes denote model predictions.

Because the ground-truth boxes are generated from segmentation masks, qualitative interpretation should consider possible annotation noise. In particular, an apparent mismatch can arise from model error, imperfect segmentation boundaries, aggregation of bowel into one gastrointestinal label, or geometric changes introduced by resizing and slice resampling. Nevertheless, the overlay is useful for checking whether the detector localizes the correct anatomical region and whether errors mainly affect organ center, in-plane extent, or depth coverage.

These qualitative samples complement the quantitative mAP results. At lower IoU thresholds, a prediction can be counted as correct when it overlaps the proper organ region, which is consistent with the visually plausible solid boxes in Figure~\ref{fig:qualitative}. At stricter thresholds, small errors in box extent along any axis reduce 3D IoU substantially. The gap between dashed and solid boxes therefore helps explain why performance remains strong at IoU 0.1 and 0.3 but drops at IoU 0.7.

\section{Discussion}

The experiments show that pseudo-text-conditioned 3D grounding is feasible for abdominal organ localization. The strongest result is coarse-to-moderate localization: AP is high at IoU 0.1, indicating that the model usually identifies the correct anatomical region. However, AP drops to 0.4309 at IoU 0.5 and 0.1552 at IoU 0.7 for the best model, showing that precise 3D box regression remains weak. This is expected for volumetric boxes because small boundary errors along multiple axes compound into a large IoU penalty, especially when supervision is derived from segmentation masks that may themselves contain residual errors.

The pretraining ablation highlights a limitation of transferring a classification-pretrained backbone to detection. Classification pretraining encourages global semantic discrimination, while detection requires spatially precise and scale-aware representations. In our pretrained variants, only the Swin3D feature extractor is initialized from a patient-level classification checkpoint; the feature enhancer, query selection module, decoder, classifier, and box regressor are still trained from scratch. The observed performance gap therefore suggests that future pretraining should be more closely matched to localization, for example through self-supervised volumetric reconstruction, segmentation-derived detection pretraining, box-supervised pretraining, or large-scale 3D multimodal pretraining.

The pseudo text design is intentionally lightweight. It is useful when no large-scale 3D language-image pretraining is available, but it limits semantic richness because the tokens omit organ names, anatomical relations, and radiology report context. Thus, the current method is best described as pseudo-text- or class-conditioned detection. This limitation directly motivates future work: language-based pretraining on larger medical datasets could improve feature extraction, class semantics, and generalization to new organs or findings.

The current evaluation focuses on organ boxes. This is a deliberate first step because reliable organ localization can support several downstream tasks, including organ-specific injury classification, attention-based report analysis, and candidate region extraction for segmentation models. However, organ localization alone leaves trauma detection unresolved. Injuries may occupy only a small part of an organ, may be subtle in intensity, or may require contrast-phase context. A future injury-focused version should use the localized organ boxes as spatial priors and add injury classification or lesion-level detection heads.

Another limitation is the absence of a comprehensive main-paper comparison against strong standard 3D detection baselines such as 3D RetinaNet, Faster R-CNN-style detectors, MONAI detection models, non-semantic 3D Swin detectors, or a purely visual 3D DETR variant. The query-selection appendix includes a module-level nnDetection reference from the supporting report, but stronger external baselines are still needed to quantify exactly how much pseudo-text conditioning contributes beyond a well-tuned closed-set 3D detector.

The 2D Grounding DINO family benefits from multi-scale deformable attention heads, which sample sparse locations across feature levels and improve convergence. Our current 3D version uses multi-scale Swin3D features and dense cross-attention as a practical first translation to volumetric data. A natural next step is to implement true 3D multi-scale deformable attention, where queries sample sparse points across depth, height, width, and scale. This could improve strict-IoU localization while reducing the cost of attention over dense 3D tokens.

Finally, the long-term goal is to move beyond fixed organ categories. Exemplar- and prototype-based medical detection has shown that semantic guidance can support robust lesion detection with fewer annotations in 2D \cite{bhat2025exemplar}. Extending this idea from 2D to 3D would allow \method{}-style detectors to use exemplars or prototypes for novel organ, lesion, or injury classes. This direction is especially relevant for 3D few-shot detection, MRI, other CT applications, and rare traumatic findings where dense annotation is expensive.

\section{Conclusion}

We presented \method, a pseudo-text-conditioned 3D Grounding DINO-style framework for organ-level 3D bounding-box localization in abdominal CT. The method adapts the semantic-conditioned query-based detection paradigm to 3D volumetric data using a Swin3D backbone, frozen random pseudo text class tokens, bidirectional feature enhancement, language-guided query selection, and a cross-modality decoder. On the limited RSNA segmentation subset, the best multi-scale model achieves 0.5830 overall mAP, with strong coarse-to-moderate localization performance and lower precision at strict 3D IoU thresholds. We release \method as an open-source baseline to encourage further development in 3D multimodal detection with minimal annotations. The main limitations are the lack of true language-pretrained 3D features, limited precise box regression, and incomplete comparison against strong standard 3D detection baselines. These limitations motivate future work on large-scale medical multimodal pretraining, true 3D multi-scale deformable attention, injury detection and classification, exemplar- or prototype-conditioned 3D detection, MRI and other CT applications, and few-shot detection for novel 3D medical findings.

\paragraph{Code availability.}
The project code is available at \url{https://github.com/SiqiChen9/3d-grounding-dino}.

\bibliographystyle{plain}
\bibliography{references}

@inproceedings{liu2024grounding,
  title={Grounding dino: Marrying dino with grounded pre-training for open-set object detection},
  author={Liu, Shilong and Zeng, Zhaoyang and Ren, Tianhe and Li, Feng and Zhang, Hao and Yang, Jie and Jiang, Qing and Li, Chunyuan and Yang, Jianwei and Su, Hang and others},
  booktitle={European conference on computer vision},
  pages={38--55},
  year={2024},
  organization={Springer}
}

@article{hermans2024rsna,
  title={Rsna 2023 abdominal trauma ai challenge: Review and outcomes},
  author={Hermans, Sebastiaan and Hu, Zixuan and Ball, Robyn L and Lin, Hui Ming and Prevedello, Luciano M and Berger, Ferco H and Yusuf, Ibrahim and Rudie, Jeffrey D and Vazirabad, Maryam and Flanders, Adam E and others},
  journal={Radiology: Artificial Intelligence},
  volume={7},
  number={1},
  pages={e240334},
  year={2024},
  publisher={Radiological Society of North America}
}

@article{rudie2024rsna,
  title={The rsna abdominal traumatic injury ct (ratic) dataset},
  author={Rudie, Jeffrey D and Lin, Hui-Ming and Ball, Robyn L and Jalal, Sabeena and Prevedello, Luciano M and Nicolaou, Savvas and Marinelli, Brett S and Flanders, Adam E and Magudia, Kirti and Shih, George and others},
  journal={Radiology: Artificial Intelligence},
  volume={6},
  number={6},
  pages={e240101},
  year={2024},
  publisher={Radiological Society of North America}
}

@inproceedings{carion2020end,
  title={End-to-end object detection with transformers},
  author={Carion, Nicolas and Massa, Francisco and Synnaeve, Gabriel and Usunier, Nicolas and Kirillov, Alexander and Zagoruyko, Sergey},
  booktitle={European conference on computer vision},
  pages={213--229},
  year={2020},
  organization={Springer}
}

@article{zhu2020deformable,
  title={Deformable detr: Deformable transformers for end-to-end object detection},
  author={Zhu, Xizhou and Su, Weijie and Lu, Lewei and Li, Bin and Wang, Xiaogang and Dai, Jifeng},
  journal={arXiv preprint arXiv:2010.04159},
  year={2020}
}

@article{zhang2022dino,
  title={Dino: Detr with improved denoising anchor boxes for end-to-end object detection},
  author={Zhang, Hao and Li, Feng and Liu, Shilong and Zhang, Lei and Su, Hang and Zhu, Jun and Ni, Lionel M and Shum, Heung-Yeung},
  journal={arXiv preprint arXiv:2203.03605},
  year={2022}
}

@article{lin2024ct,
  title={Ct-glip: 3d grounded language-image pretraining with ct scans and radiology reports for full-body scenarios},
  author={Lin, Jingyang and Xia, Yingda and Zhang, Jianpeng and Yan, Ke and Cao, Kai and Lu, Le and Luo, Jiebo and Zhang, Ling},
  journal={arXiv preprint arXiv:2404.15272},
  year={2024}
}

@article{yang2025swin3d,
  title={Swin3d: A pretrained transformer backbone for 3d indoor scene understanding},
  author={Yang, Yu-Qi and Guo, Yu-Xiao and Xiong, Jian-Yu and Liu, Yang and Pan, Hao and Wang, Peng-Shuai and Tong, Xin and Guo, Baining},
  journal={Computational Visual Media},
  volume={11},
  number={1},
  pages={83--101},
  year={2025},
  publisher={TUP}
}

@inproceedings{bhat2025exemplar,
  title={Exemplar Med-DETR: Toward Generalized and Robust Lesion Detection in Mammogram Images and Beyond},
  author={Bhat, Sheethal and Georgescu, Bogdan and Panambur, Adarsh Bhandary and Zinnen, Mathias and Nguyen, Tri-Thien and Mansoor, Awais and Elbarbary, Karim Khalifa and Bayer, Siming and Ghesu, Florin-Cristian and Grbic, Sasa and others},
  booktitle={International Conference on Medical Image Computing and Computer-Assisted Intervention},
  pages={205--215},
  year={2025},
  organization={Springer}
}

@article{antonelli2022medical,
  title={The medical segmentation decathlon},
  author={Antonelli, Michela and Reinke, Annika and Bakas, Spyridon and Farahani, Keyvan and Kopp-Schneider, Annette and Landman, Bennett A and Litjens, Geert and Menze, Bjoern and Ronneberger, Olaf and Summers, Ronald M and others},
  journal={Nature communications},
  volume={13},
  number={1},
  pages={4128},
  year={2022},
  publisher={Nature Publishing Group UK London}
}

@article{wasserthal2023totalsegmentator,
  title={TotalSegmentator: robust segmentation of 104 anatomic structures in CT images},
  author={Wasserthal, Jakob and Breit, Hanns-Christian and Meyer, Manfred T and Pradella, Maurice and Hinck, Daniel and Sauter, Alexander W and Heye, Tobias and Boll, Daniel T and Cyriac, Joshy and Yang, Shan and others},
  journal={Radiology: Artificial Intelligence},
  volume={5},
  number={5},
  pages={e230024},
  year={2023},
  publisher={Radiological Society of North America}
}

@article{rajchl2016deepcut,
  title={Deepcut: Object segmentation from bounding box annotations using convolutional neural networks},
  author={Rajchl, Martin and Lee, Matthew CH and Oktay, Ozan and Kamnitsas, Konstantinos and Passerat-Palmbach, Jonathan and Bai, Wenjia and Damodaram, Mellisa and Rutherford, Mary A and Hajnal, Joseph V and Kainz, Bernhard and others},
  journal={IEEE transactions on medical imaging},
  volume={36},
  number={2},
  pages={674--683},
  year={2016},
  publisher={IEEE}
}

@article{yan2018deeplesion,
  title={DeepLesion: automated mining of large-scale lesion annotations and universal lesion detection with deep learning},
  author={Yan, Ke and Wang, Xiaosong and Lu, Le and Summers, Ronald M},
  journal={Journal of medical imaging},
  volume={5},
  number={3},
  pages={036501--036501},
  year={2018},
  publisher={Society of Photo-Optical Instrumentation Engineers}
}

\clearpage
\appendix

\section{Swin3D Backbone Details}
\label{app:backbone}

This appendix summarizes the Swin3D backbone design developed by Han Gong for \method. The backbone provides the volumetric visual tokens used by the rest of the detector. Its role is to convert the input CT tensor into a hierarchy of local-to-global 3D representations while keeping memory consumption low enough for detector training.

\begin{figure}[H]
    \centering
    \includegraphics[width=0.88\linewidth]{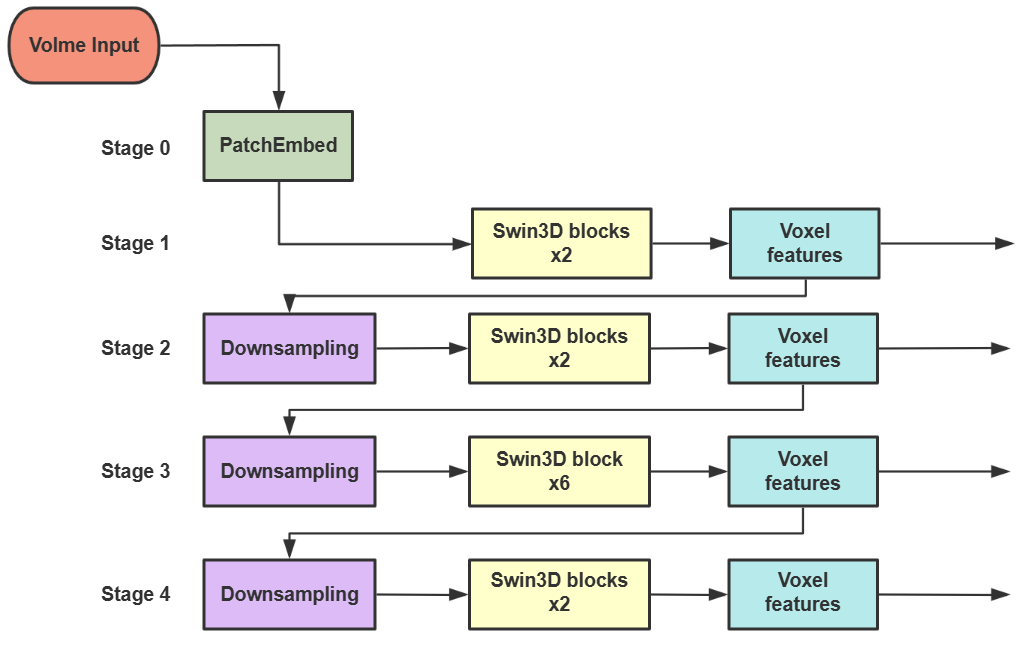}
    \caption{Swin3D backbone architecture used in the supporting backbone analysis. The CT volume is first converted into 3D patch embeddings, then processed by hierarchical Swin Transformer stages with window-based attention and patch merging. The selected stage outputs provide multi-scale volumetric features for the downstream detector.}
    \label{fig:appendix_swin3d_pipeline}
\end{figure}
\FloatBarrier

\paragraph{Patch embedding and tensor layout.}
The input volume is represented as a 5D tensor $(B,C,D,H,W)$, where $C=1$ for single-channel CT. A 3D patch embedding layer applies a convolution whose kernel size and stride match the patch size, typically $(4,4,4)$. This partitions the volume into non-overlapping cuboidal patches and projects each patch to an embedding vector. The output is first produced in channel-first form $(B,d,D',H',W')$ and is then permuted to $(B,D',H',W',d)$ so that layer normalization can be applied along the feature dimension. This step reduces the raw voxel grid to a manageable token grid while preserving local anatomical structure.

\paragraph{Window-based 3D self-attention.}
The core computational issue in volumetric transformers is that global attention grows quickly with the number of 3D tokens. The backbone therefore uses window attention: the feature map is partitioned into local 3D windows, and self-attention is computed within each window. For token $i$ and attention head $h$, the attended feature can be written as
\begin{equation}
f^{\star}_{i,h}=\sum_{j=1}^{N_w}\alpha_{ij,h} f_j W_{V,h}, \qquad
\alpha_{ij,h}=\operatorname{softmax}\left(\frac{f_i W_{Q,h}(f_j W_{K,h})^\top}{\sqrt{d_k}} + r_{ij}\right),
\end{equation}
where $N_w$ is the number of tokens in a local window and $r_{ij}$ denotes the relative position bias. Relative position bias is important because spatial relationships between CT tokens carry anatomical meaning. To exchange information across neighboring windows, alternating blocks use shifted-window attention. An attention mask prevents invalid connections across shifted window boundaries.

\paragraph{Swin Transformer stages.}
The backbone follows a hierarchical Swin-style design. Each \texttt{SwinTransformerBlock3D} contains window multi-head self-attention, a feed-forward MLP, residual connections, and layer normalization. Window attention (W-MSA) and shifted-window attention (SW-MSA) are alternated so that local features are refined while cross-window context can propagate through depth. The default stage depths are $[2,2,6,2]$, with attention heads $[3,6,12,24]$ in the corresponding stages.

\begin{figure}[H]
    \centering
    \includegraphics[width=0.70\linewidth]{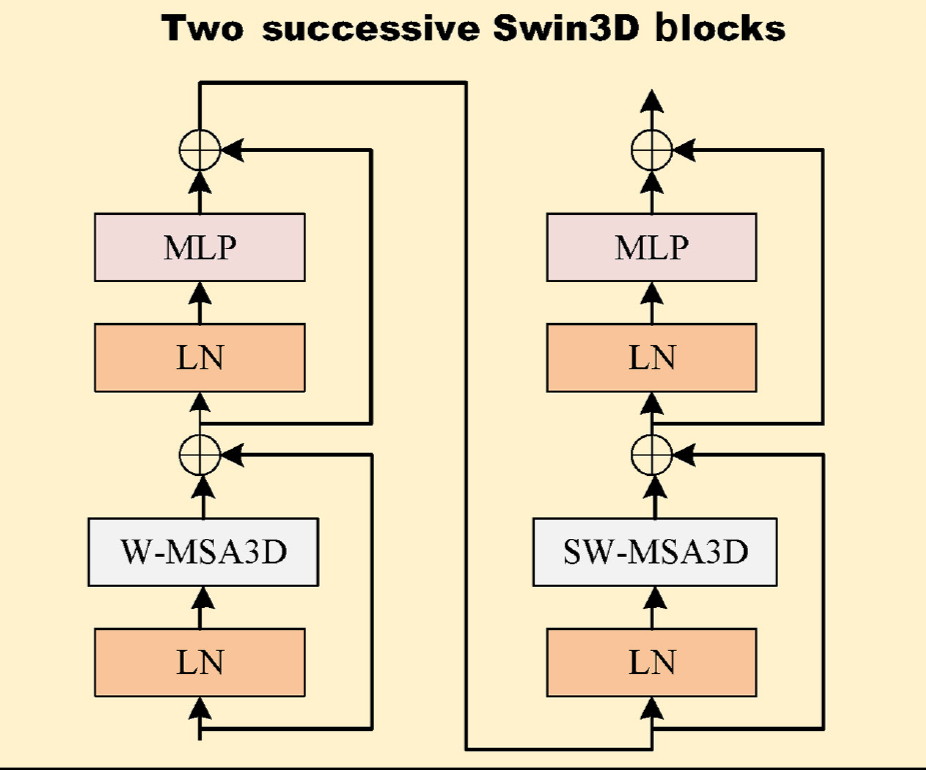}
    \caption{Swin Transformer block with window multi-head self-attention and shifted-window multi-head self-attention, reproduced from the Swin3D backbone report.}
    \label{fig:appendix_swin3d_block}
\end{figure}
\FloatBarrier

\paragraph{Patch merging and multi-scale outputs.}
Between stages, \texttt{PatchMerging3D} downsamples the spatial grid by grouping neighboring $2\times2\times2$ tokens. If the input to patch merging is $(B,D,H,W,C)$, the output has spatial size approximately $(D/2,H/2,W/2)$ and increased channel capacity. This produces progressively coarser but more semantic feature maps. The implementation supports an \texttt{out\_indices} argument, so the detector can return either the final stage or multiple selected stages. In \method, these multi-scale outputs are projected to a shared hidden dimension and flattened before being passed to the feature enhancer and query-selection modules.

\paragraph{Implementation validation.}
The backbone report validates the implementation with output-shape tests, gradient-flow tests, and small-data overfitting checks. Shape tests confirm that each stage and patch-merging operation produces the expected tensor dimensions. Gradient tests verify that the backbone can be optimized end-to-end. The overfitting check confirms that the backbone has sufficient capacity to fit a small classification subset, which is a useful sanity check before integrating it into the full detector.

\paragraph{Backbone experiments.}
The backbone report evaluates full attention and window attention under increasing target widths on an NVIDIA A100 GPU with 40 GB memory. Experiments use the RSNA 2023 data, AdamW optimization, and the detector losses used by the full framework. The first experimental phase measures memory scalability; the second compares full attention, window attention, and multi-scale window attention under detection training. The main observation is that window attention gives similar detection behavior while drastically reducing memory, allowing higher-resolution inputs that are infeasible with full attention.

\begin{table}[t]
\centering
\begin{tabular}{lccc}
\toprule
\textbf{Method} & \textbf{Width 64} & \textbf{Width 96} & \textbf{Width 128} \\
\midrule
Full attention & 31.99 & -- & -- \\
Window attention & 4.91 & 13.35 & 17.38 \\
\bottomrule
\end{tabular}
\caption{Peak memory usage in GB for full-attention and window-attention Swin3D backbones.}
\label{tab:backbone_memory}
\end{table}

The memory comparison in Table~\ref{tab:backbone_memory} shows why local attention is necessary for 3D CT. At target width 64, full attention already requires 31.99 GB, while window attention requires only 4.91 GB. Full attention runs out of memory at target widths 96 and 128, whereas window attention remains feasible at 13.35 GB and 17.38 GB. This makes window attention a practical requirement for volumetric training.

\begin{table}[t]
\centering
\begin{tabular}{lccc}
\toprule
\textbf{Method} & \textbf{mAP@0.10} & \textbf{mAP@0.50} & \textbf{mAP@0.70} \\
\midrule
full64 & 0.9734 & 0.4655 & 0.1094 \\
win64 & 0.9730 & 0.4797 & 0.0918 \\
win64-MS & 0.9764 & 0.4147 & 0.1104 \\
win128-MS & 0.9649 & 0.4309 & 0.1552 \\
\bottomrule
\end{tabular}
\caption{Backbone-related detection ablation from the Swin3D report. Window attention enables higher-resolution and multi-scale settings under memory constraints.}
\label{tab:backbone_map}
\end{table}

Table~\ref{tab:backbone_map} summarizes the detection-oriented ablation transferred from the Swin3D report. The full-attention and window-attention models perform similarly at target width 64, but only window attention can scale to target width 128. The multi-scale window model improves strict-threshold localization at mAP@0.70, which supports the design choice of using selected later-stage feature maps in the final detector.

\clearpage
\section{Bidirectional Feature Enhancer Details}
\label{app:feature_enhancer}

This appendix summarizes the feature enhancer developed by Keyi Hou for \method. The feature enhancer is the intermediate multimodal alignment module between the Swin3D visual backbone, the pseudo text token generator, the language-guided query selection module, and the cross-modality decoder. Its purpose is to avoid sending independently encoded image and text features directly to the decoder. Instead, it introduces an early stage of explicit image--text interaction so that query selection and decoding operate on already aligned features.

\begin{figure}[H]
    \centering
    \includegraphics[height=0.68\textheight,keepaspectratio]{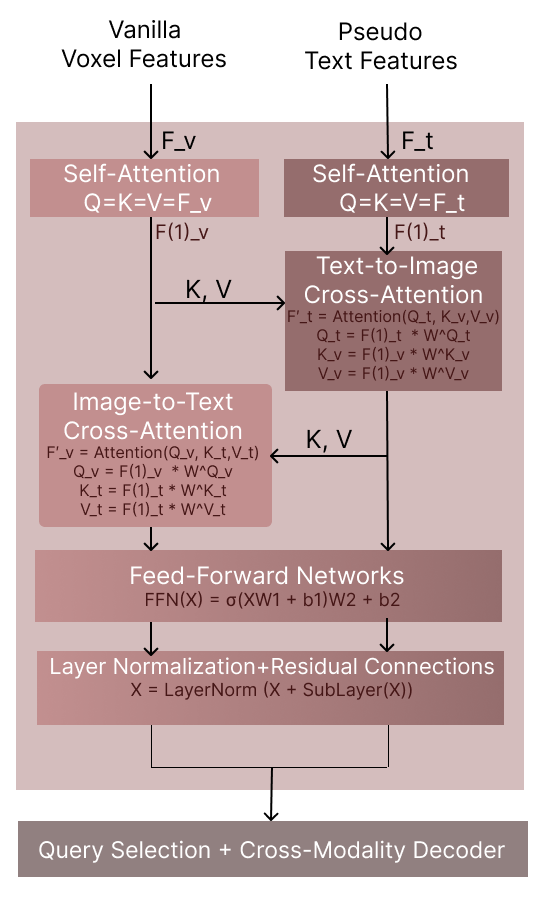}
    \caption{Bidirectional feature enhancer architecture used in the feature-enhancement analysis. Visual tokens and pseudo-text class tokens are refined with modality-specific self-attention and bidirectional cross-attention before being passed to query selection and decoding.}
    \label{fig:appendix_feature_enhancer_pipeline}
\end{figure}
\FloatBarrier

\paragraph{Motivation and inputs.}
The visual branch encodes anatomical appearance, spatial layout, and volumetric context, while the pseudo text branch encodes class-level semantics. If these streams are only fused inside the final decoder, the decoder must learn cross-modal correspondence and box prediction simultaneously. The feature enhancer addresses this by refining both modalities before query selection. It receives visual features $F_v \in \mathbb{R}^{B\times N_v\times C}$ and pseudo text features $F_t \in \mathbb{R}^{B\times N_t\times C}$, where $N_v$ and $N_t$ denote the number of visual and text tokens. It outputs enhanced visual and text features with the same dimensions, preserving compatibility with downstream modules.

\paragraph{Layer structure.}
Each feature enhancer layer contains modality-specific self-attention, bidirectional cross-attention, feed-forward networks, residual connections, and layer normalization. Self-attention first models dependencies within each modality:
\begin{equation}
F_v^{(1)}=\operatorname{SelfAttn}(F_v), \qquad F_t^{(1)}=\operatorname{SelfAttn}(F_t).
\end{equation}
The image-to-text branch uses image features as queries and text features as keys and values:
\begin{equation}
F'_v=\operatorname{Attention}(Q_v,K_t,V_t), \qquad
Q_v=F_v^{(1)}W_Q^v,\quad K_t=F_t^{(1)}W_K^t,\quad V_t=F_t^{(1)}W_V^t,
\end{equation}
while the text-to-image branch uses text features as queries and image features as keys and values:
\begin{equation}
F'_t=\operatorname{Attention}(Q_t,K_v,V_v), \qquad
Q_t=F_t^{(1)}W_Q^t,\quad K_v=F_v^{(1)}W_K^v,\quad V_v=F_v^{(1)}W_V^v.
\end{equation}
The attention function is scaled dot-product attention,
\begin{equation}
\operatorname{Attention}(Q,K,V)=\operatorname{softmax}\left(\frac{QK^\top}{\sqrt{d_k}}\right)V.
\end{equation}
After cross-attention, modality-specific feed-forward networks refine the features:
\begin{equation}
\operatorname{FFN}(X)=\sigma(XW_1+b_1)W_2+b_2.
\end{equation}
Each sub-layer is wrapped with residual connections and layer normalization,
\begin{equation}
X \leftarrow \operatorname{LayerNorm}(X + \operatorname{SubLayer}(X)),
\end{equation}
which stabilizes optimization and keeps feature scales consistent across stacked layers.

\paragraph{Interpretation of the two directions.}
The image-to-text direction conditions visual tokens on organ semantics, which is especially important because the final box regression depends directly on visual feature quality. The text-to-image direction grounds class tokens in the CT context, improving semantic representations before they are used by query selection and the decoder. The report argues that the two directions play complementary roles with different effects: removing image-to-text attention causes the larger performance drop because visual features lose direct semantic guidance.

\paragraph{Implementation validation.}
The feature enhancer report includes several unit tests. Shape tests verify that text and image tensors preserve their expected dimensions after enhancement, ensuring that the output can be passed directly to query selection and decoding. Spatial-size tests vary the flattened visual token count $N_v$ over values such as 32, 64, 128, and 256, confirming that the module is robust to different feature-map resolutions. Feature-transformation tests compare inputs and outputs to confirm that the module performs a non-identity transformation. Gradient-flow tests backpropagate a simple loss through the module and verify that gradients reach the input features and parameters. Additional tests cover configurable depth, such as one, two, or three stacked layers, and deterministic evaluation-mode behavior.

\paragraph{Experimental setup.}
The ablations are conducted inside the full 3D Grounding-DETR framework on RSNA 2023 CT volumes. All compared configurations use the same backbone, pseudo text generator, query selection, decoder, training schedule, batch size, and data augmentation. The feature-enhancer report studies four settings: a baseline with no feature enhancer, the full bidirectional feature enhancer, a variant without image-to-text cross-attention, and a variant without text-to-image cross-attention.

\paragraph{Training behavior.}
The report observes that the baseline without the feature enhancer can reduce training loss, but its validation loss remains higher and more unstable in later epochs. With the feature enhancer, validation behavior is smoother and localization-related losses decrease more consistently. Gradient norms are also more controlled, suggesting that early multimodal interaction makes the optimization problem better conditioned. These trends support the idea that the module improves generalization beyond a simple increase in model capacity.

\begin{figure}[H]
    \centering
    \includegraphics[width=0.82\linewidth]{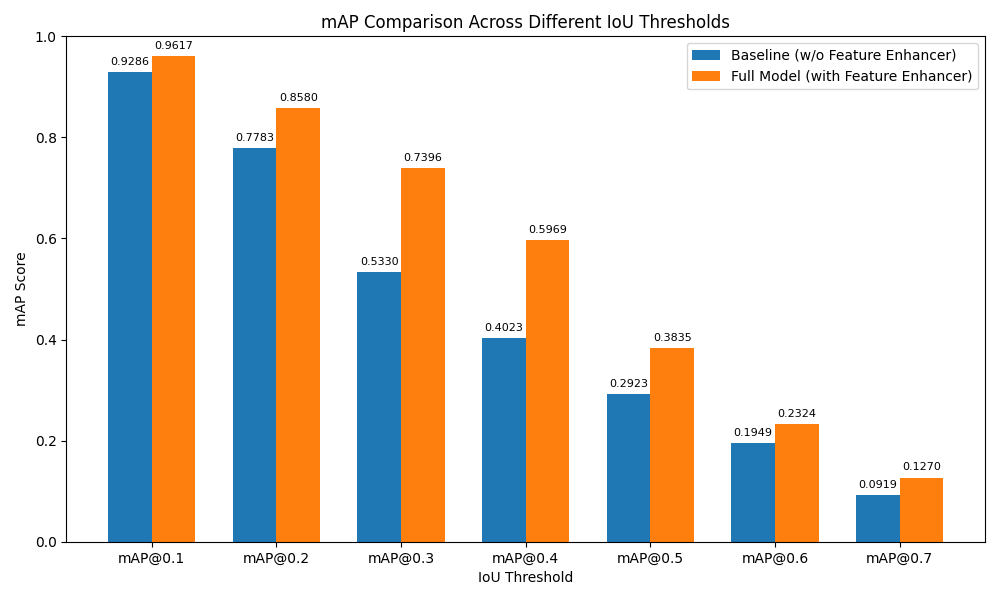}
    \caption{mAP comparison with and without the feature enhancer, reproduced from the feature-enhancement report.}
    \label{fig:appendix_feature_enhancer_map}
\end{figure}

\paragraph{Test-set performance.}
The feature-enhancement report presents the test-set mAP comparison as plots rather than as tables. Figure~\ref{fig:appendix_feature_enhancer_map} reproduces the report's comparison between the baseline without the feature enhancer and the full model with the proposed module. The report states that the full feature enhancer improves overall mAP from 0.4602 to 0.5570, an absolute gain of 0.0968. The largest improvements appear at intermediate thresholds: mAP@0.3 increases from 0.5330 to 0.7396, and mAP@0.4 increases from 0.4023 to 0.5969. This indicates that the module improves both coarse detection and spatial alignment quality.

\FloatBarrier

\begin{figure}[H]
    \centering
    \includegraphics[width=0.82\linewidth]{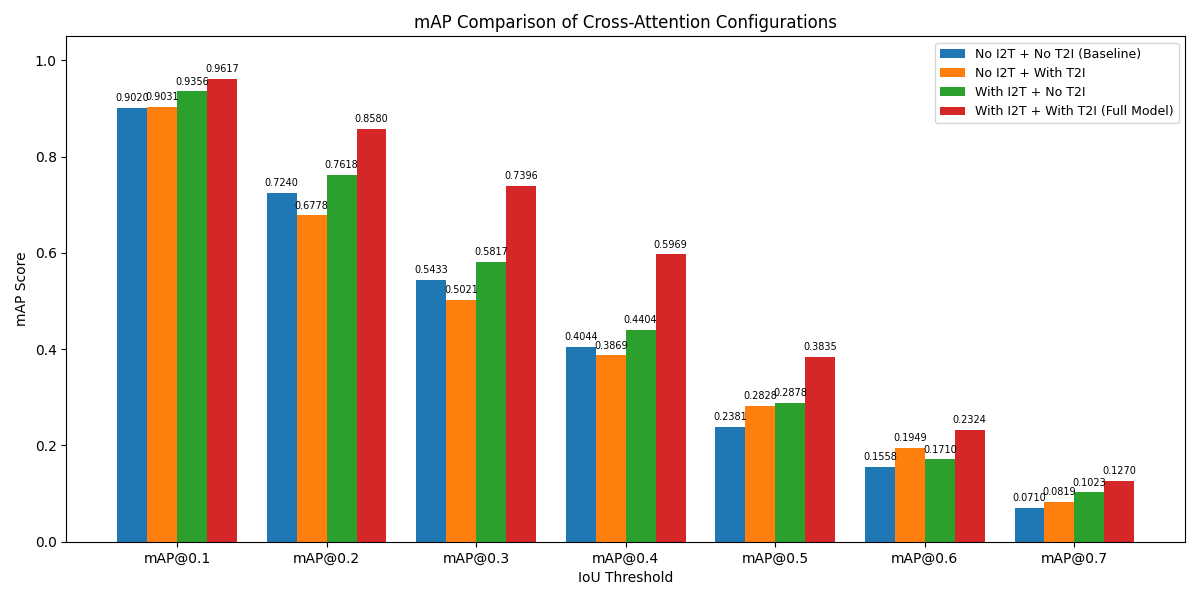}
    \caption{mAP comparison of feature-enhancer cross-attention configurations, reproduced from the feature-enhancement report.}
    \label{fig:appendix_feature_enhancer_direction_map}
\end{figure}

\paragraph{Cross-attention ablation.}
The report also visualizes the directional cross-attention ablation in Figure~\ref{fig:appendix_feature_enhancer_direction_map}. Removing text-to-image attention leaves the image-to-text branch active and obtains 0.4687 overall mAP, whereas removing image-to-text attention obtains 0.4328. The same ordering appears at mAP@0.3 and mAP@0.4. This supports the report's conclusion that semantic conditioning of the visual stream is particularly important for precise 3D grounding.

\FloatBarrier

\clearpage
\section{Language-Guided Query Selection Details}
\label{app:query_selection}

This appendix summarizes the language-guided query selection module developed by Jingxuan Yang for \method. The module sits between the feature enhancer and the cross-modality decoder. Its goal is to initialize decoder queries from image tokens that are already semantically aligned with organ-class pseudo text tokens.

\begin{figure}[H]
    \centering
    \includegraphics[width=0.88\linewidth,height=0.68\textheight,keepaspectratio]{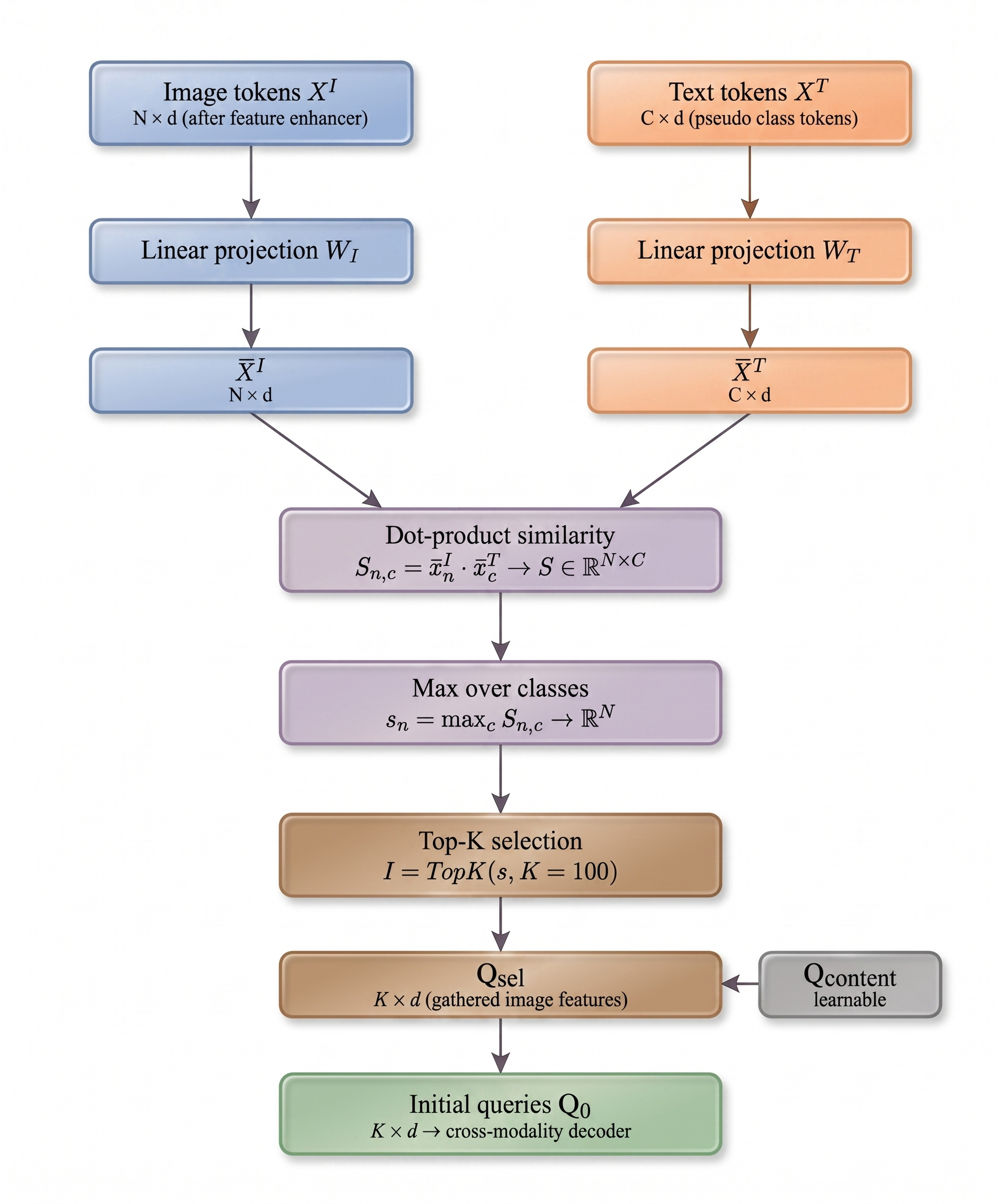}
    \caption{Language-guided query selection architecture used in the query-selection analysis. Enhanced image tokens and pseudo-text tokens are projected into a shared space, scored by semantic relevance, and ranked to initialize decoder queries from the most relevant visual locations.}
    \label{fig:appendix_query_selection_flow}
\end{figure}
\FloatBarrier

\paragraph{Motivation.}
Standard DETR-style detectors use a fixed set of learned queries that must discover object regions through decoder training. This is inefficient for 3D CT because the number of visual tokens is large and the target objects occupy only part of the volume. The query-selection report follows the Grounding DINO intuition that query initialization should be conditioned on text or class semantics. In \method, the text branch uses pseudo class tokens, and semantic query seeding is performed by measuring visual-token affinity to these class tokens.

\paragraph{Projection to a shared space.}
Let the feature enhancer output image tokens $X^I\in\mathbb{R}^{B\times N_I\times d}$ and pseudo text tokens $X^T\in\mathbb{R}^{B\times C\times d}$. The query selection module first maps both modalities into a common latent space:
\begin{equation}
\bar{X}^I = W_I X^I, \qquad \bar{X}^T = W_T X^T,
\end{equation}
where $W_I$ and $W_T$ are learnable linear projections. This step reduces distribution mismatch between dense visual features and semantic class tokens before similarity computation.

\paragraph{Similarity and relevance scoring.}
A token--class affinity matrix is computed by dot product:
\begin{equation}
S= \bar{X}^I (\bar{X}^T)^\top, \qquad S\in\mathbb{R}^{B\times N_I\times C}.
\end{equation}
For each image token, the maximum similarity across the class dimension is used as the semantic relevance score:
\begin{equation}
s_n = \max_{c} S_{n,c}.
\end{equation}
The maximum operator preserves tokens that strongly match at least one class. This is useful when target structures are localized and class frequencies are imbalanced, because mean aggregation could dilute a strong response for a small or rare structure.

\paragraph{Top-$K$ query construction.}
The indices of the $K$ highest-scoring visual tokens are selected dynamically:
\begin{equation}
\mathcal{I}=\operatorname{TopK}(s,K), \qquad Q_{\mathrm{sel}}=\operatorname{Gather}(X^I,\mathcal{I}).
\end{equation}
The selected visual tokens are then combined with learnable content embeddings:
\begin{equation}
Q_0 = Q_{\mathrm{sel}} + Q_{\mathrm{content}}.
\end{equation}
The learnable content term helps break symmetry among selected tokens and gives the decoder trainable query capacity. If fewer than $K$ visual tokens are available, the implementation pads the selected tensor so that the decoder always receives a fixed query shape.

\paragraph{Design trade-off.}
The query budget $K$ controls the balance between spatial coverage and background noise. A small $K$ can miss relevant regions, especially if multiple organs compete for query slots. A very large $K$ admits many redundant or background tokens, which can make Hungarian matching less stable. The module report therefore evaluates several values of $K$ and finds that $K=100$ is the strongest setting in its ablation protocol.

\paragraph{Implementation validation.}
The query-selection report includes an eight-stage unit-test suite. The tests verify output tensor shape $(K,B,d)$ under varying token counts, hyperparameter flexibility for $K\in\{50,100,200\}$, batch behavior for $B\in\{1,2,4,8\}$, compatibility with feature dimensions such as $d\in\{256,512,768\}$, gradient flow through the image and text projections, sensitivity to different pseudo-text matrices, default FP32 precision, and correct behavior when $K$ exceeds the total number of available visual tokens. These checks confirm that the module can be inserted into the full detector without breaking tensor contracts or gradient propagation.

\paragraph{Module-level comparison.}
Table~\ref{tab:query_init_appendix} reports the query-selection ablation. These numbers are retained as a module-level analysis of query initialization and should be interpreted separately from the final main-paper comparison in Section~\ref{sec:final_quantitative} and Figure~\ref{fig:map_grouped}, which use the final multi-scale setting and report the final overall mAP of 0.5830.

\begin{table}[t]
\centering
\small
\setlength{\tabcolsep}{3pt}
\begin{tabular}{llccccc}
\toprule
\textbf{Method} & \textbf{Variant} & \textbf{mAP@0.1} & \textbf{mAP@0.3} & \textbf{mAP@0.5} & \textbf{mAP@0.7} & \textbf{Mean} \\
\midrule
nnDetection$^{\dagger}$ & U-Net based 3D anchor & 0.912 & 0.642 & 0.354 & 0.182 & 0.523 \\
CT-3GDINO & Learnable queries only & 0.892 & 0.585 & 0.320 & 0.150 & 0.487 \\
CT-3GDINO & Random initialization & 0.934 & 0.665 & 0.375 & 0.224 & 0.549 \\
CT-3GDINO & Language-guided selection & 0.958 & 0.712 & 0.468 & 0.315 & 0.613 \\
\bottomrule
\end{tabular}
\caption{Updated module-level comparison from the query-selection report. $^{\dagger}$The nnDetection row is reported by the supporting query-selection report as an external reference on the same RSNA 2023 test split; the CT-3GDINO rows isolate query initialization while sharing the same backbone and decoder.}
\label{tab:query_init_appendix}
\end{table}

The updated report finds that visual-token initialization is better than purely learned queries, and language-guided visual-token selection is best among the compared CT-3GDINO options. The largest relative benefit appears at stricter IoU thresholds: language-guided selection reaches 0.315 mAP@0.7, compared with 0.224 for random initialization, 0.150 for learnable-only queries, and 0.182 for the nnDetection reference. This suggests that semantically meaningful query seeds help the decoder refine boundaries and detect coarse organ locations.

\begin{table}[t]
\centering
\small
\begin{tabular}{lccccc}
\toprule
\textbf{Selected queries $K$} & \textbf{mAP@0.1} & \textbf{mAP@0.3} & \textbf{mAP@0.5} & \textbf{mAP@0.7} & \textbf{Mean} \\
\midrule
25 & 0.908 & 0.641 & 0.318 & 0.148 & 0.504 \\
50 & 0.919 & 0.663 & 0.332 & 0.157 & 0.518 \\
100 & 0.958 & 0.712 & 0.468 & 0.315 & 0.613 \\
200 & 0.890 & 0.576 & 0.308 & 0.135 & 0.477 \\
\bottomrule
\end{tabular}
\caption{Query budget sensitivity from the query-selection report. The best result is obtained with $K=100$, indicating a coverage--noise trade-off.}
\label{tab:query_k_appendix}
\end{table}

Table~\ref{tab:query_k_appendix} shows a non-monotonic dependence on $K$. Increasing from 25 to 100 improves the mean score, but increasing further to 200 degrades performance. This supports the interpretation that too few queries under-cover candidate organ regions, while too many queries introduce redundant background tokens into the decoder and matching process.

\begin{figure}[H]
    \centering
    \includegraphics[width=0.70\linewidth]{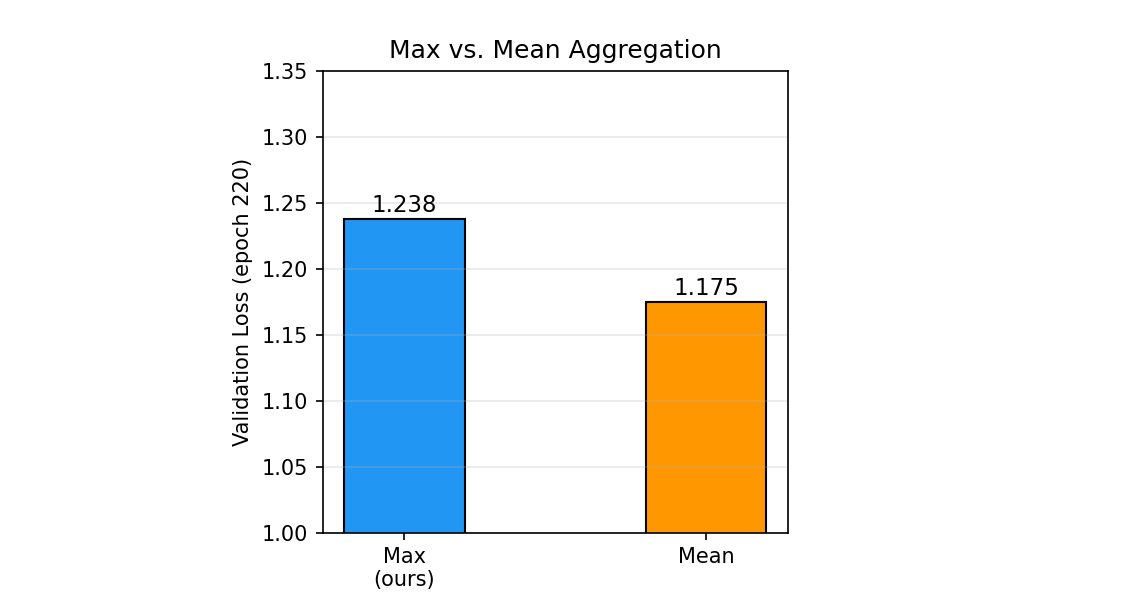}
    \caption{Validation loss for max and mean aggregation strategies, reproduced from the updated query-selection source.}
    \label{fig:query_agg_appendix}
\end{figure}

\paragraph{Aggregation strategy.}
The updated query-selection source also compares max aggregation with mean aggregation across class scores. Figure~\ref{fig:query_agg_appendix} reproduces the validation-loss plot from the report. At epoch 220, max aggregation gives a validation loss of 1.238, while mean aggregation gives 1.175. The report interprets the two curves as broadly comparable and keeps max aggregation as the default because it preserves the strongest class-specific response for each image token, which is useful when localized or rare categories could be diluted by mean pooling.

\FloatBarrier

\begin{figure}[H]
    \centering
    \includegraphics[width=0.92\linewidth]{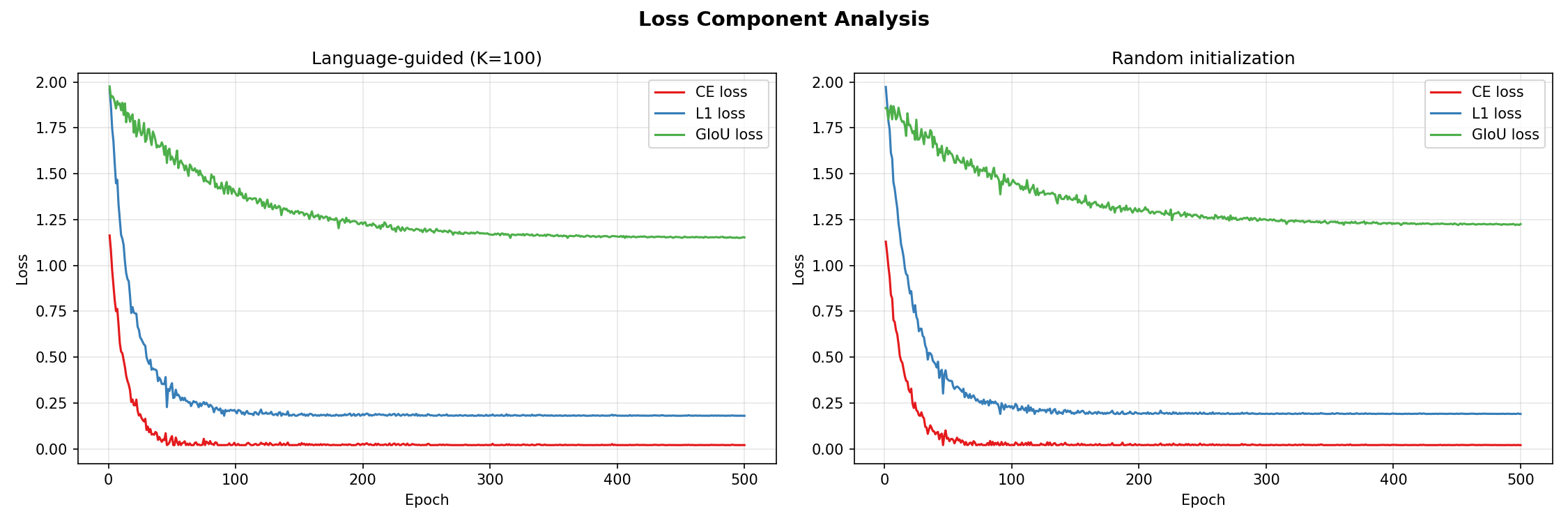}
    \caption{Loss component analysis for language-guided query selection and random initialization over 500 epochs, reproduced from the query-selection report.}
    \label{fig:query_loss_components_appendix}
\end{figure}

\begin{figure}[H]
    \centering
    \includegraphics[width=0.86\linewidth]{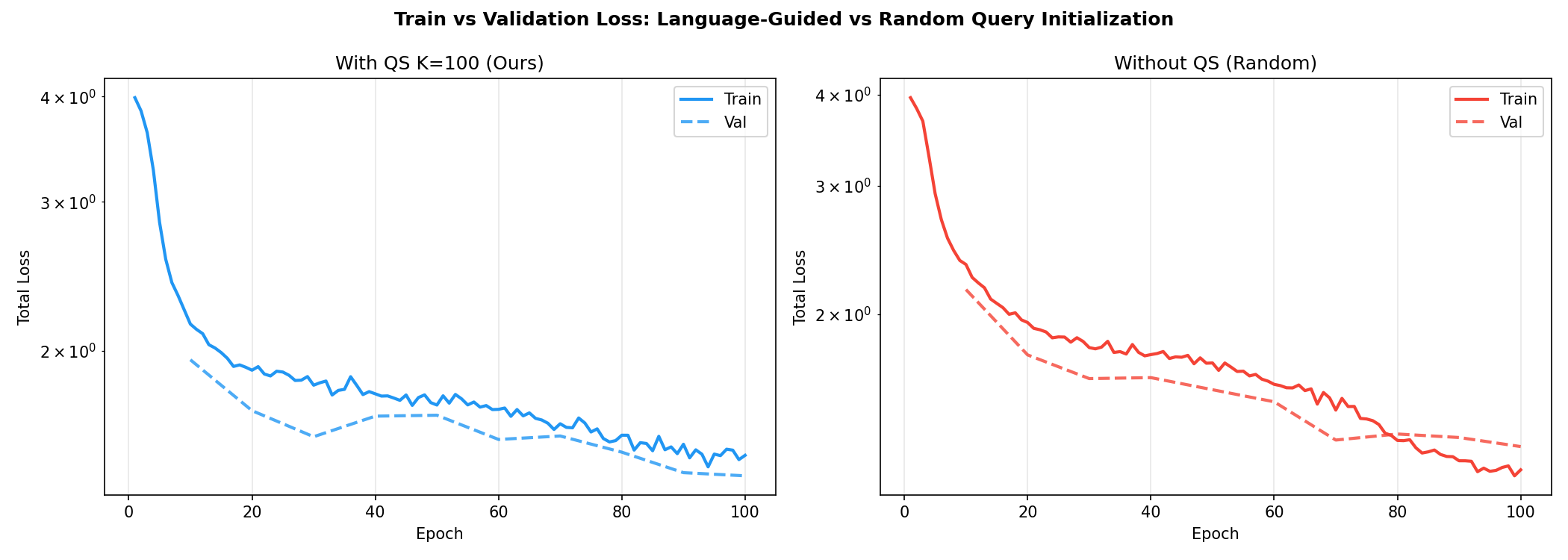}
    \caption{Training and validation loss curves for language-guided query selection and random initialization over 500 epochs, reproduced from the query-selection report.}
    \label{fig:query_train_val_appendix}
\end{figure}

\paragraph{Optimization behavior.}
The query-selection report includes training diagnostics for the language-guided and random-initialization variants. Figure~\ref{fig:query_loss_components_appendix} shows the classification, L1, and GIoU loss components over 500 epochs. The report identifies GIoU as the slowest-converging term, indicating that 3D boundary regression is the main optimization bottleneck. Figure~\ref{fig:query_train_val_appendix} compares train and validation losses and shows that validation loss tracks training loss closely, suggesting no obvious overfitting in this module-level experiment.

\FloatBarrier

\FloatBarrier

\begin{figure}[H]
    \centering
    \includegraphics[width=0.88\linewidth]{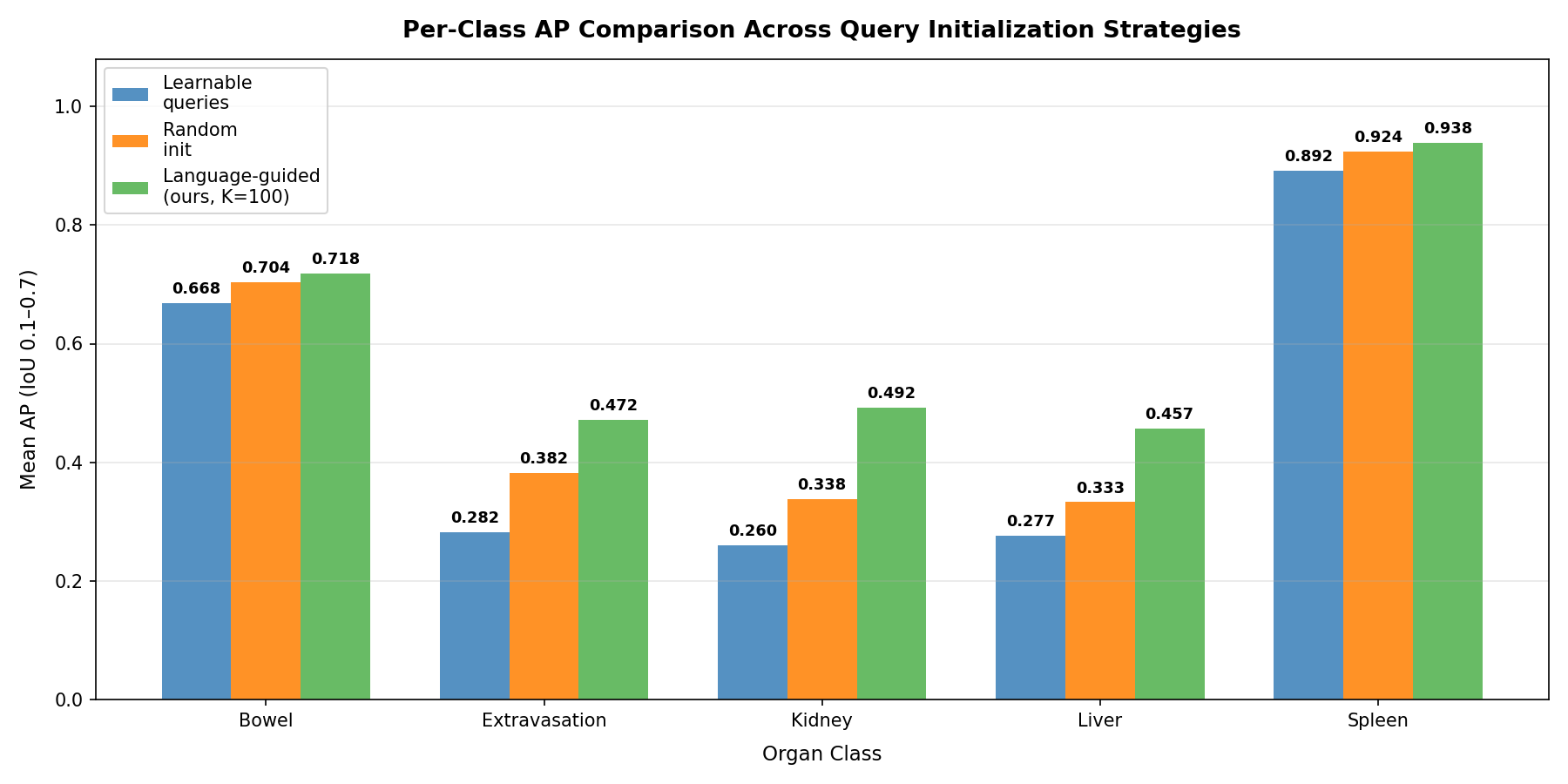}
    \caption{Per-class mean AP comparison across query-initialization strategies, reproduced from the query-selection report.}
    \label{fig:query_perclass_appendix}
\end{figure}

\paragraph{Per-class diagnostic analysis.}
The query-selection report also provides the per-class diagnostic breakdown in Table~\ref{tab:query_perclass_appendix} and Figure~\ref{fig:query_perclass_appendix}. This table is retained as a module-level diagnostic from the supporting report. Its category grouping follows the query-selection report and is therefore not identical to the final five-label organ-localization vocabulary used in the main paper, where the kidneys are separated into left and right kidney and extravasation is not a target organ class.

\begin{table}[t]
\centering
\small
\begin{tabular}{lccccc}
\toprule
\textbf{Report category} & \textbf{AP@0.1} & \textbf{AP@0.3} & \textbf{AP@0.5} & \textbf{AP@0.7} & \textbf{Mean} \\
\midrule
Bowel & 1.000 & 0.912 & 0.745 & 0.215 & 0.718 \\
Extravasation & 1.000 & 0.620 & 0.185 & 0.082 & 0.472 \\
Kidney & 0.892 & 0.584 & 0.295 & 0.198 & 0.492 \\
Liver & 0.942 & 0.512 & 0.228 & 0.145 & 0.457 \\
Spleen & 1.000 & 1.000 & 0.985 & 0.765 & 0.938 \\
\midrule
Overall & 0.958 & 0.712 & 0.468 & 0.315 & 0.613 \\
\bottomrule
\end{tabular}
\caption{Per-class AP breakdown for the language-guided model with $K=100$, reproduced from the query-selection report as a module-level diagnostic.}
\label{tab:query_perclass_appendix}
\end{table}

\FloatBarrier

\begin{figure}[H]
    \centering
    \includegraphics[width=0.92\linewidth]{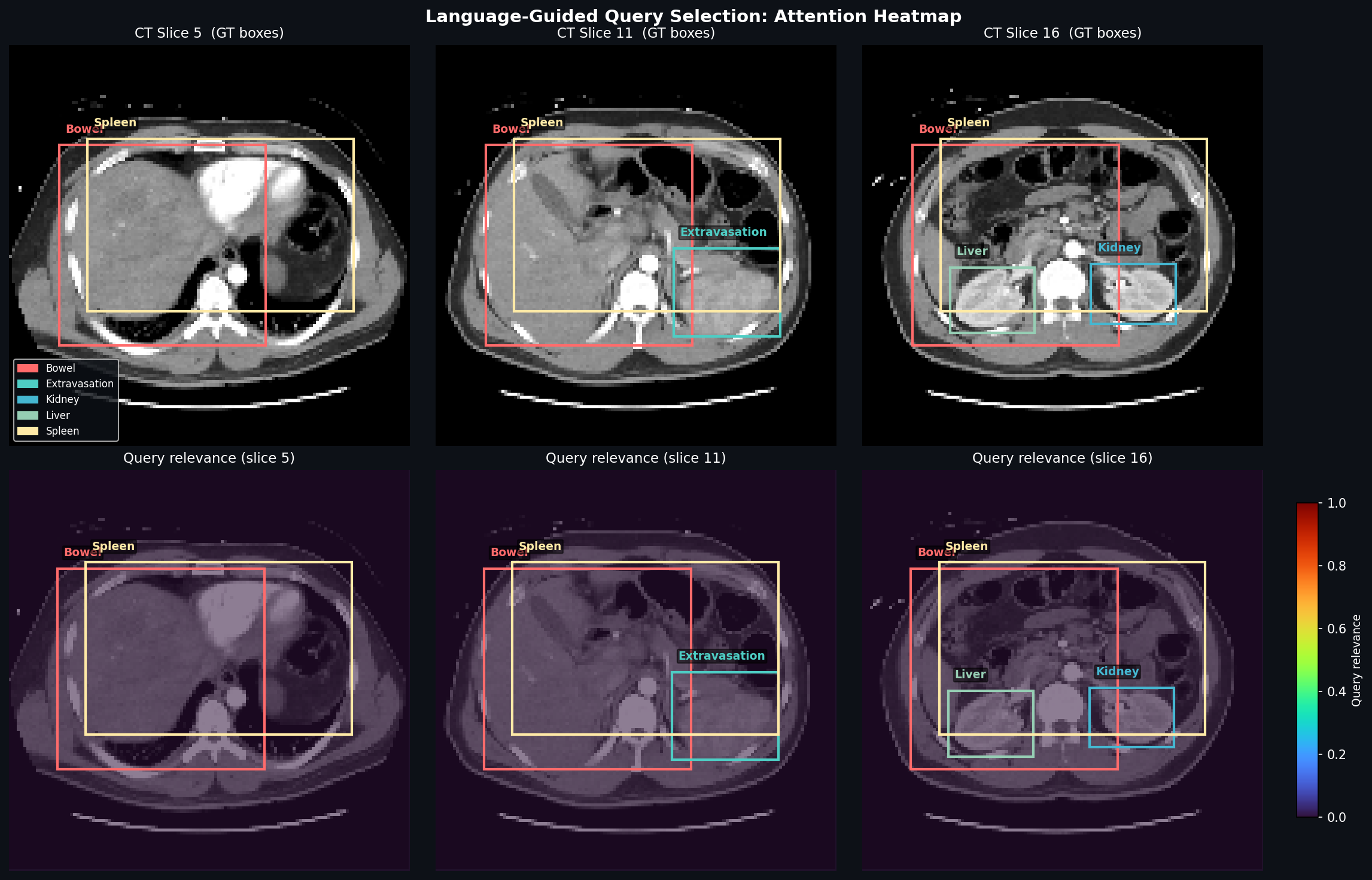}
    \caption{Query-selection attention heatmap with ground-truth boxes, reproduced from the query-selection report.}
    \label{fig:query_vis_appendix}
\end{figure}

\begin{figure}[H]
    \centering
    \includegraphics[width=0.86\linewidth]{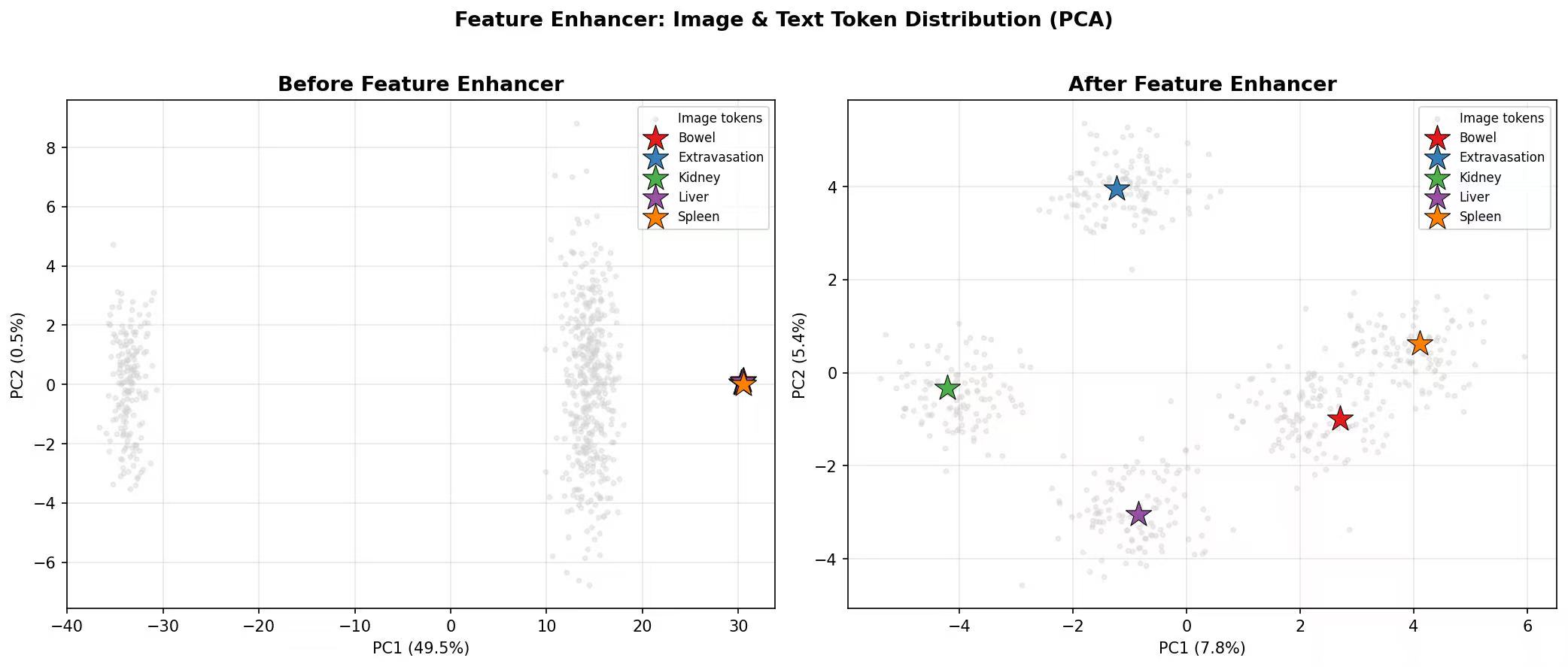}
    \caption{PCA of image tokens and pseudo-text tokens before and after feature enhancement, reproduced from the query-selection report.}
    \label{fig:query_pca_appendix}
\end{figure}

\begin{figure}[H]
    \centering
    \includegraphics[width=0.92\linewidth]{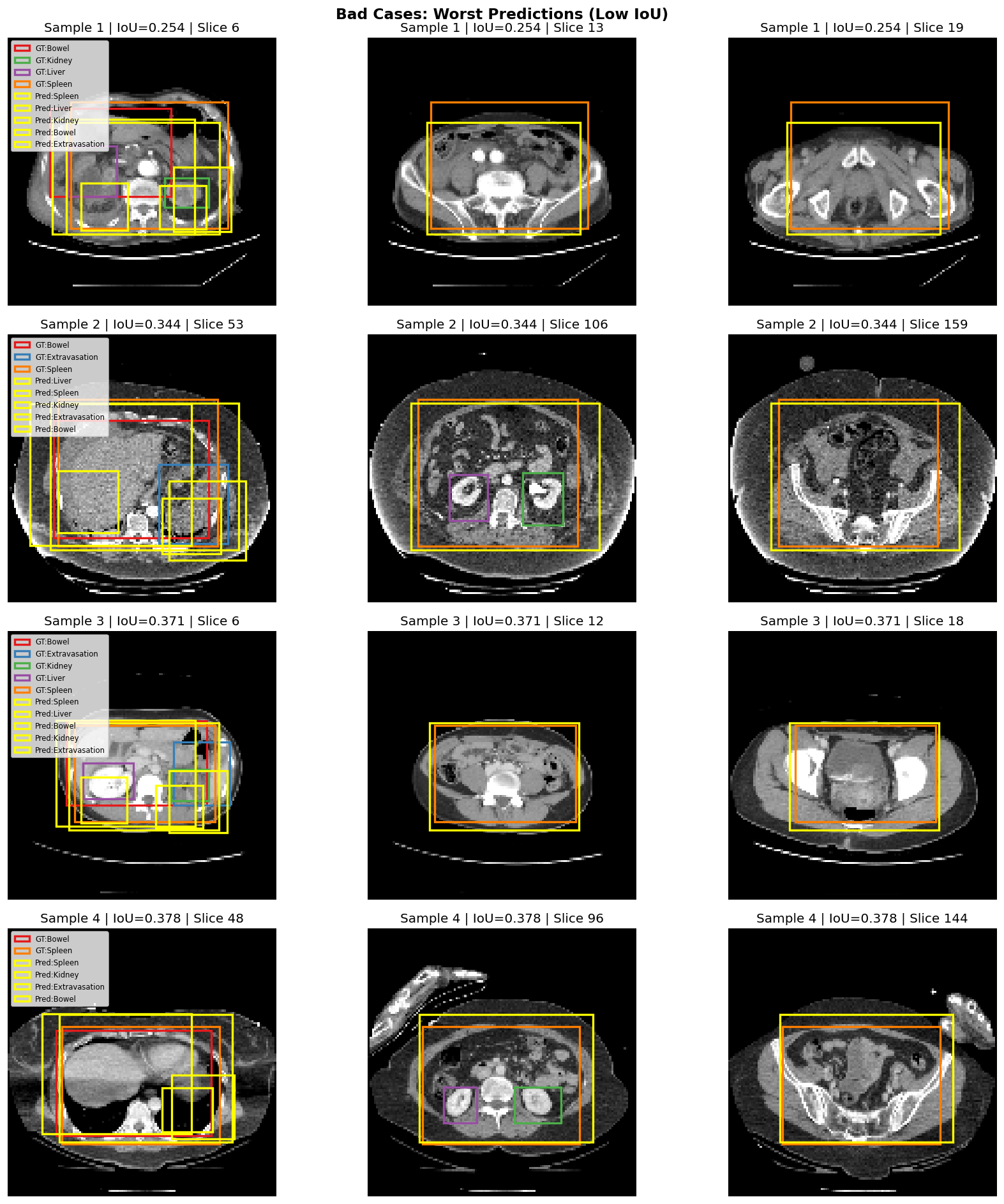}
    \caption{Worst-prediction examples from the query-selection report. Ground truth is shown in color and predictions are shown in yellow; the dominant failure mode is over-large boxes.}
    \label{fig:query_bad_cases_appendix}
\end{figure}

\paragraph{Visual diagnostics.}
Figure~\ref{fig:query_vis_appendix} reproduces the query-selection heatmap from the report. High-response regions concentrate around the central abdomen and overlap plausible target regions, supporting the intended role of pseudo-text affinity as a query-ranking signal. Figure~\ref{fig:query_pca_appendix} shows the report's PCA analysis before and after feature enhancement, where pseudo-text tokens become better separated from the dense image-token cloud after enhancement. Figure~\ref{fig:query_bad_cases_appendix} shows the worst predictions identified in the report, where the dominant failure mode is over-large predicted boxes. These qualitative diagnostics are consistent with the quantitative pattern: coarse localization is strong, while strict IoU remains limited by boundary over-extension.

\FloatBarrier

\FloatBarrier

\FloatBarrier

\paragraph{Limitations and possible extensions.}
The current relevance score uses a maximum over classes. This preserves strong class-specific responses but can favor large or frequent structures if their tokens dominate the top-$K$ set. Future versions could use class-balanced Top-$K$ selection, organ-specific query budgets, diversity constraints, or uncertainty-aware query sampling. These extensions would keep the same interface to the decoder while improving coverage of smaller or more difficult organs.

\end{document}